\documentclass[journal]{IEEEtran}
\usepackage{tabularx,booktabs}
\usepackage{multirow}  
\usepackage{ifpdf}
\usepackage{bbding} 
\usepackage{enumerate}
 \ifpdf
 \else
 \fi

\usepackage{cite}

\ifCLASSINFOpdf
   \usepackage[pdftex]{graphicx}
\else
\fi

\usepackage{amsmath}

\usepackage{algorithmic}
\usepackage{algorithm}

\usepackage{array}

\ifCLASSOPTIONcompsoc
  \usepackage[caption=false,font=normalsize,labelfont=sf,textfont=sf]{subfig}
\else
  \usepackage[caption=false,font=footnotesize]{subfig}
\fi

\usepackage{url}

\makeatletter
\let\NAT@parse\undefined
\makeatother
\usepackage{hyperref}  

\usepackage{lipsum}
\makeatletter
\newenvironment{breakablealgorithm}
{
	\begin{center}
		\refstepcounter{algorithm}
		\hrule height.8pt depth0pt \kern2pt
		\renewcommand{\caption}[2][\relax]{
			{\raggedright\textbf{\ALG@name~\thealgorithm} ##2\par}%
			\ifx\relax##1\relax 
			\addcontentsline{loa}{algorithm}{\protect\numberline{\thealgorithm}##2}%
			\else 
			\addcontentsline{loa}{algorithm}{\protect\numberline{\thealgorithm}##1}%
			\fi
			\kern2pt\hrule\kern2pt
		}
	}{
		\kern2pt\hrule\relax
	\end{center}
}
\makeatother

\begin{document}

\title{When does the Physarum Solver Distinguish the Shortest Path from other Paths: the Transition Point and its Applications}

\author{Yusheng~Huang, Dong~Chu, Joel~Weijia~Lai, Yong~Deng, Kang~Hao~Cheong
\thanks{Y. Huang, D. Chu and Y. Deng are with the Institute of Fundamental and Frontier Science, University of Electronic Science and Technology of China, Chengdu, 610054, China (e-mail: dengentropy@uestc.edu.cn, prof.deng@hotmail.com). }
\thanks{J.W. Lai and K.H. Cheong is with Science and Math Cluster, Singapore University of Technology and Design (SUTD), S487372, Singapore (e-mail:kanghao\_cheong@sutd.edu.sg).}
}

\markboth{IEEE TRANSACTIONS}%
{Huang \MakeLowercase{\textit{et al.}}:The Capacity Constraint Physarum Solver}
\ifCLASSOPTIONpeerreview
\fi

\maketitle
\begin{abstract}
Physarum solver, also called the physarum polycephalum inspired algorithm (PPA), is a newly developed bio-inspired algorithm that has an inherent ability to find the shortest path in a given graph. Recent research has proposed methods to develop this algorithm further by accelerating the original PPA (OPPA)'s path-finding process. However, when does the PPA ascertain that the shortest path has been found? Is there a point after which the PPA could distinguish the shortest path from other paths? By innovatively proposing the concept of the dominant path (D-Path), the exact moment, named the transition point (T-Point), when the PPA finds the shortest path can be identified. Based on the D-Path and T-Point, a newly accelerated PPA named OPPA-D using the proposed termination criterion is developed which is superior to all other baseline algorithms according to the experiments conducted in this paper. The validity and the superiority of the proposed termination criterion is also demonstrated. Furthermore, an evaluation method is proposed to provide new insights for the comparison of different accelerated OPPAs. The breakthrough of this paper lies in using D-path and T-point to terminate the OPPA. The novel termination criterion reveals the actual performance of this OPPA. This OPPA is the fastest algorithm, outperforming some so-called accelerated OPPAs. Furthermore, we explain why some existing works inappropriately claim to be accelerated algorithms is in fact a product of inappropriate termination criterion, thus giving rise to the illusion that the method is accelerated.
\end{abstract}


\IEEEpeerreviewmaketitle

\section{Introduction}
\label{sec.1}
Bio-inspired algorithm (BA) has attracted attention and intrigued many due to its wide range of applications for a long time. Numerous kinds of BAs, such as the differential evolution algorithm\cite{A_DE}, grey wolf optimizer \cite{A_GW}, artificial bee colony algorithm \cite{A_ABC}, genetic algorithm \cite{A_GA}, ant colony algorithm (ACO)\cite{A_ACO}, and whale optimization algorithm \cite{A_WO} have been developed. BAs have played important roles in solving optimization problems, such as large-scale multiobjective detection and optimization \cite{Zhang2017,Tian2020}, parameter estimation \cite{A_example4}, global numerical optimization \cite{A_example1}, feature selection \cite{A_example3}, numerical function optimization \cite{A_ABC}, and boosting support vector machine \cite{A_example2}, just to name a few. In the past decade, a new BA, named the physarum polycephalum inspired algorithm (PPA) \cite{GG_PPA,GG_PPA2}, also called the physarum solver, has attracted great attention.

The physarum polycephalum, a kind of slime mold, is a single-celled amoeboid organism \cite{GG_PPA}. In 2000, Nakagaki \textit{et al.} \cite{A_PPAmaze} observed that the amoeboid organism would gradually change its shape to improve its efficiency of foraging, which leads to the result that its body would only cover the shortest path in the mazes after some time. The maze-solving ability of the slime mold was then considered as a kind of primitive intelligence \cite{A_PPAsmart}. It is believed that the maze-solving mechanism which is also the path-finding mechanism was related to the contraction waves in the organism \cite{A_PPApath}. The contraction waves influence the thickness of the tube, resulting in a positive mechanism. If the flow through a certain tube persists or increases for a certain period, the tube will become thicker; otherwise, the tube will contract \cite{GG_IPPA}.

In 2007, a mathematical model of the physarum polycephalum, which is the original physarum polycephalum inspired algorithm (OPPA) was proposed \cite{GG_PPA2}. In the OPPA, the tube-liked body of the physarum polycephalum is considered as a network with the growing point and the food source serving as the starting point and ending point, respectively; the Poiseuille flow is adopted to model the flow flowing through the tube; the conservation law of flux (also called continuity of flux) is obeyed using network Poisson equation \cite{GG_PPA}. At the core of the OPPA is the use of a novel adaptive equation to model the dynamics of tube thickness. By using the techniques mentioned above, the OPPA perfectly models the path-finding behavior of the amoeboid organism and was then applied to solve the minimum-risk path-finding problems \cite{GG_PPA2} and adaptive network design problems \cite{GG_PPA}. The effectiveness of the OPPA has been mathematically proven. Bonifaci \textit{et al.} provided a mathematical proof of the convergence of the OPPA on the shortest path problems \cite{A_prove2} and provided the time complexity bounds \cite{A_prove1}. Karrenbauer \textit{et al.} developed on the former proof and proposed a more general proof for the effectiveness of the OPPA \cite{A_prove3}.

Many relevant research have been proposed after the initial introduction of the OPPA. Some studies adopt the PPA to improve the performance of other BAs. For example, PPA is utilized to initialize the ant colony algorithm to solve some NP-Hard problems \cite{A_PPAex_1}; Gao \textit{et al.} combined some traditional BAs with PPA in network community detection; a combined PPA framework with the genetic algorithm is proposed to handle the traveling salesman problem \cite{A_PPAex_3}, just to name a few. Other studies aim to expand the PPA's applications. For example, Sun \textit{et al.} \cite{A_PPAex_4} proposed a physarum-inspired approach to identify sub-networks for drug re-positioning; Xu \textit{et al.} \cite{A_PPAex_5} developed a multi-sink-multi-source PPA to tackle the traffic assignment problem; Tsompanas \textit{et al.} \cite{Tsompanas2015}, and Zhang \textit{et al.} \cite{Zhang2018} also worked on finding equilibrium state of traffic assignments using PPA; Jiang \textit{et al.} \cite{A_PPAex_6} applied the PPA in routing protocol design; Song \textit{et al.} \cite{A_PPAex_7} designed a PPA-based optimizer for minimum exposure problem, etc. 

Besides the research mentioned above, some researchers are also interested in improving the convergence rate of the OPPA. The computational complexity for solving the network Poisson equation using PPA is $O(n^3)$. While it is polynomial time, the typical number of variables is of a much higher order, thus resulting in an inefficient allocation of time resource before converging to the shortest path \cite{GG_IPPA}. Thus, several studies related to accelerated OPPA have been developed. Zhang \textit{et al.} \cite{GG_IPPA} proposed an improved OPPA by introducing the concept of energy. In the improved PPA, it is assumed that the foraging of the physarum polycephalum consumes energy and the law of the conservation of energy should be satisfied. In another work, the OPPA is accelerated by eliminating some near vanished tubes of the OPPA \cite{GG_EHPA}. Wang \textit{et al.} \cite{GG_ANPA} believed that the tubes in the shortest path would be more competitive than others when OPPA converges. Thus they proposed an anticipation mechanism to terminate the OPPA earlier. Cai \textit{et al.} \cite{GG_BPPA} combined the OPPA with the Bayesian rule to achieve a higher convergence rate and also proposed a method to tackle negative-weighted edges in the OPPA. Gao \textit{et al.} \cite{GG_APS} developed an accelerated OPPA by excluding the inactive nodes and collecting the near-optimal paths. The above researchers share the same motivation---while the OPPA is effective, its efficiency can be greatly improved; hence accelerating the OPPA is necessary. These work have sped up the OPPA using different methods. Yet, despite significant achievements being made by previous studies, there remain some questions that need to be answered:
\begin{enumerate}[1)]
	\item \textbf{In the case of ensuring accuracy, is there an inherent limit so that no matter how the termination criteria changes, the OPPA cannot increase the convergence rate?} Both \cite{GG_ANPA} and \cite{GG_APS} stated the necessity of defining an appropriate stopping condition for the OPPA and claimed that the proposed solutions could terminate the OPPA earlier. This leads to a secondary question, how much earlier? Is there a limit for the early termination, prior to which the OPPA have not converged to the shortest path?
	\item \textbf{Is there a better evaluation method that could more intuitively evaluate how much improvement the accelerated methods are making?} All of the studies mentioned above used either the running time or the number of iterations to compare different accelerated OPPAs. However, the two metrics are highly dependent on the chosen stopping criteria, in other words, if the evaluation method is dependent on the stopping criteria, the comparison may not be objective. Hence, is there an evaluation method that is not dependent to the stopping criteria of the OPPA?
\end{enumerate}
The two questions above ultimately lead to the main question:\textbf{ when does the OPPA distinguish the shortest path from the other paths?}  

The contributions of this work are listed below:
\begin{itemize}
	\item We have defined two concepts, i.e., the dominant path and the transition point of the OPPA. By combining the use of these two concepts, we identify the exact moment that the OPPA starts to detect the shortest path for the first time. These concepts answer the first and main questions.
	\item Combining the defined concepts, we propose a new stopping criterion for the OPPA. By combining the proposed criterion and the OPPA, a new algorithm named the OPPA-D is developed. According to the experimental results in Section \ref{sec.4.1}, the OPPA is the fastest among all of the tested algorithms. The validity and the superiority of the proposed criterion are also demonstrated in Section \ref{sec.4.2}.
	\item By adopting the concept of the transition point, we have developed a novel method to compare different accelerated OPPAs more objectively. By doing so, the second question can be answered. Using the proposed methods, we compare our algorithm with existing accelerated OPPAs in Section \ref{sec.4.3} where we also present some key findings.
\end{itemize}

Broadly, the paper is developed as follows: Section \ref{sec.2} introduces the background and development of the PPA; Section \ref{sec.3} provides the defined concepts and the proposed algorithm; the proposed algorithm is then experimentally tested in Section \ref{sec.4}; Section \ref{sec.4} also provides the proposed evaluation method and the corresponding analysis. Section \ref{sec.5} discusses and concludes the paper.

\section{Brief introduction of the original PPA}
\label{sec.2}
This section is mainly based on \cite{GG_PPA,GG_PPA2}, where more details could be found. The pseudo-code of the OPPA is provided in Section \ref{sec.2.2} for the readers' reference.
\subsection{The mathematical model of the physarum polycephalum}
\label{sec.2.1}
As mentioned, the slime mold's body would form a tube-like network when foraging. Consider the network as a graph $G(N,E)$ where $N$ is the set of nodes and $E$ represents the set of edges. Note that $G$ is assumed to be an undirected connected graph without cycles and negative-weight edges, except where specifically mentioned. Other abstract mappings are as follows: the tubes of the network are treated as edges in $G(N,E)$; the junctions between the tube are regarded as nodes in $G(N,E)$; the growing point of the slime mold is mapped to the ending node of the $G(N,E)$ while the food source is the starting node. For simplicity, in this paper, the scenario with one starting node and one ending node is considered. For scenarios with multiple ending nodes please refer to \cite{A_PPAsupply} and as for scenarios with multiple starting nodes please refer to \cite{A_PPAex_5}.

In this paper, $N_1$ and $N_2$ denotes starting and the ending nodes, respectively; the other nodes are labeled as $N_3, N_4, N_5, \cdots$; $E_{ij}$ represents the edge between nodes $N_i$ and $N_j$; $Q_{ij}$ represents the flux flowing through the edge $E_{ij}$ from node $N_i$ to node $N_j$.

In \cite{GG_PPA2}, the flow in the slime mold's network is regarded as laminar. Thus, the Hagen-Poiseuille equation is observed, giving us
\begin{equation}
	Q_{ij}=
	\frac{D_{ij}}{L_{ij}}(p_i-p_j),
	\label{eq.flowdefinition}
\end{equation}
where $p_i$ denotes the pressure at node $N_i$; $D_{ij}$ represents the conductivity of the edge $E_{ij}$ and $L_{ij}$ is the length of the edge.

The whole network is driven by the inflow at the food source (the starting node). Thus, at $N_1$, the source term of flux is
\begin{equation}
	\sum_{i\not=1}{Q_{i1}=IN_0} 
	\label{eq.inflow}
\end{equation}
where $IN_0$ is the inflow of the network.

The outflow of the network is considered equal to the inflow which is $IN_0$. Then, at the ending node ($N_2$) of the graph, we have
\begin{equation}
	\sum_{i\not=2}{Q_{i2}=-IN_0}
	\label{eq.outflow}
\end{equation}

The inflow and outflow at other nodes should be balanced, that is
\begin{equation}
	\sum_{j\not=1,2; i\not=j}{Q_{ij}=0}.
	\label{eq.balanceflow}
\end{equation}

By combining (\ref{eq.flowdefinition}) to (\ref{eq.balanceflow}), the network Poisson equation of the graph $G(N,E)$ is
\begin{align}
	\sum_{i\not=j}{\frac{D_{ij}}{L_{ij}}(p_i-p_j)}=\begin{cases}
			+IN_0 & \text{for}\ j=1,  \\
			-IN_0 & \text{for}\ j=2,  \\
			0 & \text{otherwise}.
		\end{cases}
	\label{eq.Poissonequation}
\end{align}

All $p_i$ can be calculated by solving the network Poisson equation, i.e., (\ref{eq.Poissonequation}), by setting $p_1$ to $0$. After obtaining $p_i$,  $Q_{ij}$ can be determined using (\ref{eq.flowdefinition}). However, we still need $D_{ij}$ to solve (\ref{eq.Poissonequation}). We now discuss the method to update $D_{ij}$ at each iteration which is the main feature of the OPPA.

In \cite{GG_PPA,GG_PPA2}, the following adaptation equation is adopted to model the dynamics of the thickness of tubes:
\begin{equation}
	\frac{d}{dt}D_{ij}=f(|Q_{ij}|)-\alpha D_{ij},
	\label{eq.AdaptEq}
\end{equation}
where $f(|Q_{ij}|)=|Q_{ij}|$ and $\alpha=1$ are typically used. Further discussion on different parameter settings of the adaptation equation can be found in previous literature \cite{GG_PPA2,A_prove3}. This adaptation equation suggests that the conductivity of tube tends to decrease exponentially while it increases linearly with flux along this tube.

In order to implement the adaptation equation, we discretize (\ref{eq.AdaptEq}) by performing linear approximations to get
\begin{equation}
	\frac{D_{ij}^{n+1}-D_{ij}^{n}}{\Delta t}=|Q_{ij}^n|-D_{ij}^{n+1}, 
	\label{eq.Discretize}
\end{equation}
where $Q_{ij}^n$ is the flow flowing through edge $E_{ij}$ at  $n$th iteration; $\Delta t$ is normally considered as $1$ (please refer to \cite{GG_APS} for various other considerations); $D_{ij}^{n}$ and $D_{ij}^{n+1}$ represents the conductivity of edge $E_{ij}$ at $n$-th and $(n+1)$-th iteration, respectively. A more concise form of the above formula is
\begin{equation}
	D_{ij}^{n+1}=\frac{|Q_{ij}^n|+D_{ij}^n}{2}. 
	\label{eq.7}
\end{equation}

\subsection{The pseudo-code of the OPPA}
\label{sec.2.2}
Having introduced the mathematical model for the physarum solver, we will now provide the pseudo-code of the OPPA. For the rest of the paper, `OPPA' will refer to the following algorithm which uses the original PPA to solve the shortest path problem. The pseudo-code of the OPPA is demonstrated in Alg.\ref{alg.OPPA}.

\begin{breakablealgorithm}
	\caption{The OPPA \cite{GG_PPA}}
	\label{alg.OPPA}
	\begin{algorithmic}[1]
		\begin{footnotesize}
			\STATE {//Initialization part}
			\STATE {\textbf{Input:} the statistics of graph $G(N,E)$ and the corresponding matrix of the weight of each edges $\textbf{W}$.}
			\STATE {$D_{ij}\Leftarrow0.5\ (\forall E_{ij}\in E,)$} // Initialize the conductivity of each edge.
			\STATE {$Q_{ij}\Leftarrow0\ (\forall E_{ij}\in E)$} // Initialize the flow of each edge.
			\STATE {$L_{ij}\Leftarrow W_{ij} (\forall E_{ij}\in E)$} //The length of the edges equal to their weights.
			\STATE {$p_i\Leftarrow0\ (\forall i=1,2,\cdots,N)$} // Initialize the pressure at each node.
			\STATE {$count\Leftarrow1$} // Initialize the counting variable.
			\STATE {//Iterative part}  
			\REPEAT 
			\STATE  $p_1\Leftarrow0$ // The pressure of the starting node (Node $1$) is set to 0.
			\STATE  Calculate the pressure of all the nodes using (\ref{eq.Poissonequation}):
			\begin{center}
				\begin{equation*}            
					\sum_{i\in N}{\frac{D_{ij}}{L_{ij}}(p_i-p_j)=\left \{\begin{gathered}
							+IN_0\ \ for\ j=1,  \\
							-IN_0\ \ for\ j=2,  \\
							0\ \ \ otherwise.  \\
						\end{gathered} \right.}                
				\end{equation*}
			\end{center}
			\STATE  Calculate the flux using (\ref{eq.flowdefinition}):
			\begin{center}
				$Q_{ij}\Leftarrow D_{ij}\cdot(p_i-p_j)/L_{ij}.$
			\end{center}
			\STATE  Calculate the conductivity of the iteration using (\ref{eq.7}):
			\begin{center}
				\begin{equation*}
					D_{ij}^{n+1}=\frac{|Q_{ij}^n|+D_{ij}^n}{2}. 
				\end{equation*}
			\end{center}
			\STATE  {$count\Leftarrow count+1$} 
			\UNTIL{The given termination criterion is met.}
			\STATE {\textbf{Output:} The flow matrix $\textbf{Q}$.}
		\end{footnotesize}
	\end{algorithmic}
\end{breakablealgorithm}

Given a graph, after the execution of the OPPA, according to the flow matrix $\textbf{Q}$, one would find a path that contains the most amount of inflow in the graph, this is the shortest path.

\section{Defining concepts and the proposed OPPA-D algorithm}
\label{sec.3}
\subsection{The dominant path}
\label{sec.3.1}
\noindent\textbf{Dominant path (D-Path):} Given a flow matrix generated by the PPA-based algorithms, the D-path is the path found by the following iterative procedures:
\begin{enumerate}[i.]
	\item Set up a node list $List_{Node}$ initialized as an empty list. The starting node $N_1$ of the graph would be the first element of $List_{Node}$.
	\item To find the next element of $List_{Node}$ in the graph, find all the nodes and their corresponding edges that are connect to the current last element of $List_{Node}$.
	\item Among the edges found in the second step, find the edge that contains the maximum flow. Since one of the two nodes that relate to this edge is the last element of the node set $List_{Node}$, set the other node as the next element of $List_{Node}$. 
	\item Delete the current last element of $List_{Node}$ and its corresponding edges from the graph. Push the next element of $List_{Node}$ into the list $List_{Node}$. This element would be the last element of $List_{Node}$ in the next iteration.
	\item Repeat Step ii to Step iv until the next element we found in Step iv is the ending node of the graph. The nodes in $List_{Node}$ form one of the paths in the graph. This path is defined as the \textbf{D-Path}.
\end{enumerate}

Given a flow matrix generated by the PPA-based algorithms, the D-Path can be generated following the above procedures, and the length $L_{D-Path}$ of the path can be calculated. 

\subsection{The transition point of the physarum solver}
\label{sec.3.2}
With the definition of D-Path, it is interesting to note that after the convergence of the OPPA, the algorithm will direct most (or even all) of the inflow to the shortest path; thus, when the OPPA converges, the D-Path generated is exactly the shortest path of the graph, and the $L_{D-Path}$ will be equal to the length of the shortest path. One can also deduce that the process of the OPPA finding the shortest path is indeed the process of OPPA gradually converging the D-Path to the shortest path.

However, the means to quantify the convergence of the OPPA is still an open problem. The convergence of the OPPA is typically regarded as the iteration when the conductivity of each edge in the graph does not change with significantly with the subsequent iterations. Thus, the stopping (or convergence or termination) criterion is usually set to $\sum_{i}\sum_{j\neq i}D_{ij}\leq\epsilon$. In \cite{GG_APS} $\epsilon$ is set to $10^{-5}$ while in \cite{GG_IPPA,GG_EHPA} it is set to $10^{-2}$. There is still no conclusive method or consensus for determining the parameter $\epsilon$.

In most of the current studies, we allow the algorithm to run until it achieves the pre-determined stopping criterion. This would often result in long computational time; yet it remains a widely implemented method for deciding whether the OPPA finds the shortest path. The question remains, \textbf{does the OPPA need such a long computational time to achieve convergence?} 

With our definition of D-Path, the answer is \textbf{no}. The OPPA distinguishes the shortest path before the stopping criterion is met. To demonstrate this, four complete graphs, i.e., 'Instance 1' to 'Instance 4', are randomly generated with five nodes and the length of edges varying from 1 to 10000. The OPPA is then implemented in the four instances to find the shortest path. The termination criterion is set to $\sum_{i}\sum_{j\neq i}D_{ij}\leq 10^{-3}$. At each iteration, the $L_{D-Path}$ is recorded.

\begin{figure}[htb!]
	\centering
	\includegraphics[width=0.5\textwidth]{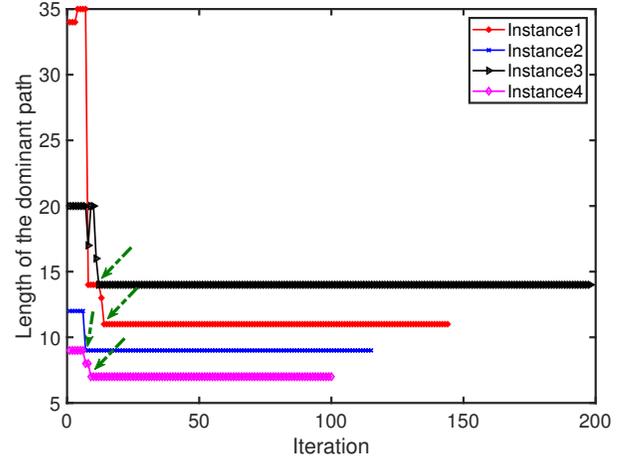}
	\caption{The transition point.}
	\label{fig.transitionpoint}
\end{figure}

Fig.\ref{fig.transitionpoint} demonstrates the results. According to Fig.\ref{fig.transitionpoint}, in all of the instances, the $L_{D-Path}$ fluctuates at the first few iterations, and then decreases rapidly, after which it converges to the length of the shortest path. Of importance is that the convergence of $L_{D-Path}$ happens long before the OPPA's termination. Thus, from Fig.\ref{fig.transitionpoint}, one could tell that the OPPA distinguishes the shortest path long before the termination condition is met.

\noindent\textbf{Transition point (T-Point):} The T-point of the PPA-based algorithms is defined as the iteration when the $L_{D-Path}$ starts to converge to the length of the shortest path. For example, the four points with green arrows pointing at in Fig.\ref{fig.transitionpoint} are the T-Points of the OPPA in each of the four instances, respectively. 

Given the definition of T-Point, the main question proposed in Section \ref{sec.1}, i.e., ``When does the OPPA distinguish the shortest path from the other paths?", can be answered. We would like to provide our opinion: after the transition point, the PPA-based algorithms start to distinguish the shortest path from the others. For the first question in Section \ref{sec.1}, i.e., ``In the case of ensuring accuracy, is there an inherent limit, so that no matter how the termination criteria changes, the OPPA cannot increase the convergence rate?", again, we would like to propose a possible answer: the inherent limit would be the T-Point, before which the OPPA does not achieve the shortest path.

\subsection{The proposed algorithm: OPPA-D}
\label{sec.3.3}
Further developing from the previous sub-sections, we now introduce our application of the proposed concepts, i.e., the OPPA-D algorithm. 

The OPPA-D algorithm builds on the OPPA with a newly proposed termination criterion. This criterion is defined as follows: Execute the OPPA until the $L_{D-Path}$ does not change for $K$ iteration. $K$ is a pre-defined parameter. Thus, after the $L_{D-Path}$ remains steady for $K$ iterations, the occurrence of the T-Point is assumed, which also implies the convergence of the OPPA.

The pseudo-code of the OPPA-D algorithm is provided in Alg.\ref{alg.OPPA-D}. For purpose of illustration, $K$ is set to $10$. The algorithm's sensitivity to $K$ will be subsequently tested in Section \ref{sec.4.2}.

\begin{breakablealgorithm}
	\caption{The OPPA-D}
	\label{alg.OPPA-D}
	\begin{algorithmic}[1]
		\begin{footnotesize}
			\STATE {//Initialization part}
			\STATE {\textbf{Input:} the statistics of graph $G(N,E)$, the corresponding matrix of the weight of each edges $\textbf{W}$, and the pre-defined parameter $K$.}
			\STATE $D_{ij}\Leftarrow0.5\ (\forall E_{ij}\in E,)$ // Initialize the conductivity of each edge.
			\STATE $Q_{ij}\Leftarrow0\ (\forall E_{ij}\in E)$// Initialize the flow of each edge.
			\STATE $L_{ij}\Leftarrow W_{ij} (\forall E_{ij}\in E)$ //The length of the edges equal to their weights.
			\STATE $p_i\Leftarrow0\ (\forall i=1,2,\cdots,N)$ // Initialize the pressure at each node.
			\STATE $count\Leftarrow1$ // Initialize the counting variable.
			\STATE $Length_{D-Path}^0\Leftarrow0$ // Variable used to record the length of D-Path.
			\STATE $COUNT\Leftarrow0$ // Variable used to count the number of iterations that the $Length_{D-Path}$ remains unchanged.
			\STATE {//Iterative part}  
			\WHILE{\textbf{true}}
			\STATE  $p_1\Leftarrow0$ // The pressure of the starting node (Node $1$) is set to 0.
			\STATE  Calculate the pressure of all the nodes using (\ref{eq.Poissonequation}):
			\begin{center}
				\begin{equation*}            
					\sum_{i\in N}{\frac{D_{ij}}{L_{ij}}(p_i-p_j)=\left \{\begin{gathered}
							+IN_0\ \ for\ j=1,  \\
							-IN_0\ \ for\ j=2,  \\
							0\ \ \ otherwise.  \\
						\end{gathered} \right.}                
				\end{equation*}
			\end{center}
			\STATE  Calculate the flux using (\ref{eq.flowdefinition}):
			\begin{center}
				$Q_{ij}\Leftarrow D_{ij}\cdot(p_i-p_j)/L_{ij}.$
			\end{center}
			\STATE  Calculate the conductivity of the iteration using (\ref{eq.7}):
			\begin{center}
				\begin{equation*}
					D_{ij}^{n+1}=\frac{|Q_{ij}^n|+D_{ij}^n}{2}. 
				\end{equation*}
			\end{center}
			\STATE Find the D-Path and calculate its length $Length_{D-Path}^{count}$ according to Section \ref{sec.3.1}.
			\IF{$Length_{D-Path}^{count}$==$Length_{D-Path}^{count-1}$}
			\STATE{$COUNT=COUNT+1$}
			\IF{$COUNT\geq K$}
			\STATE{\textbf{break}}
			\ENDIF
			\ELSE
			\STATE{$COUNT=0$}
			\ENDIF
			\STATE  {$count\Leftarrow count+1$} 
			\ENDWHILE
			\STATE {\textbf{Output:} The shortest path length $Length_{D-Path}^{count}$.}
		\end{footnotesize}
	\end{algorithmic}
\end{breakablealgorithm}

\section{Experiments \& Analysis}
\label{sec.4}
In this section, we present our experimental methodologies and our findings. In particular, we compare the proposed OPPA-D with other state-of-the-art accelerated OPPAs in randomly generated complete networks, randomly generated small-world networks, and real-world networks, to demonstrate its effectiveness and efficiency in Section \ref{sec.4.1}. Since the OPPA-D is a modified algorithm of the OPPA with the newly proposed termination criterion, we will verify the validation of the proposed stopping criterion. Hence, the OPPA-D is further compared with other commonly adopted criteria in Section \ref{sec.4.2}. In Section \ref{sec.4.3}, three accelerated OPPAs will be evaluated using a new transition-point based evaluation method. All the experiences are done through computer simulations using Matlab R2018a on an Intel Core i5-8500 CPU (3GHz) with 8 GB RAM under Windows 10.

\subsection{Comparing the OPPA-D with other accelerated OPPAs}
\label{sec.4.1}
\noindent\textbf{Data Set 1:} There are three data sets. 1) `Da-Com-1' to `Da-Com-8': Eight complete graphs are randomly generated with size ranging from 100 to 5000, and length of the edges varying from 1 to 10000. 2) `Da-SW-1' to `Da-SW-8': Eight small-world networks are randomly generated with size ranging from 100 to 5000, and length of the edges varying from 1 to 10000. All the small-world networks are created using a Matlab function 'WattsStrogatz' \footnote{More details of the function 'WattsStrogatz' could refer to https://ww2.mathworks.cn/help/matlab/math/build-watts-strogatz-small-world-graph-model.html?lang=en.}. 3) `Da-R-1' and `Da-R-2': Two real-world networks, i.e., the Sioux-Falls network and the Anaheim network \footnote{The meta-data of networks could refer to https://github.com/bstabler/TransportationNetworks.}, are adopted. Some basic information about the data sets is provided in Table \ref{table.dataset1}. 
\begin{table}
	\centering
	\caption{Basic information of Data Set 1}
	\resizebox{0.5\textwidth}{10mm}{
		\begin{tabular}{ccccccccc}
			\toprule[0.75pt]
			Instance Name&$|N|^a$&$|E|^a$&Instance Name&$|N|^a$&Mean node degree&Instance Name&$|N|^a$&$|E|^a$\\ \hline
			Da-Com-1	&	50&	1.22E+03&	Da-SW-1	&	50&	6&	Da-R-1$^b$	&	24&	76\\
			Da-Com-2	&	100&	4.95E+03&	Da-SW-2	&	100&	12&	Da-R-2$^b$	&	416&	914\\
			Da-Com-3	&	250&	3.11E+04&	Da-SW-3	&	250&	30&		&	&	\\
			Da-Com-4	&	500&	1.25E+05&	Da-SW-4	&	500&	60&		&	&	\\
			Da-Com-5	&	750&	2.81E+05&	Da-SW-5	&	750&	90&		&	&	\\
			Da-Com-6	&	1000&	4.99E+05&	Da-SW-6	&	1000&	120&		&	&	\\
			Da-Com-7	&	2000&	2.00E+06&	Da-SW-7	&	2000&	240&		&	&	\\
			Da-Com-8	&	5000&	1.25E+07&	Da-SW-8	&	5000&	600&		&	&	\\
			\bottomrule[0.75pt] 
		\end{tabular}
	}
	\\\footnotesize{$^a$$N$ is the number of nodes and $E$ is the number of edges.} \\\footnotesize{$^b$Instances `Da-R-1' and `Da-R-2'are the Sioux-Falls network and the Anaheim network, respectively. }
	\label{table.dataset1}
\end{table} 

\noindent\textbf{Baseline algorithms:} The OPPAs with termination criteria $\epsilon\leq10^{-2}$ and $\epsilon\leq10^{-5}$, i.e., the OPPA($10^{-2}$) and OPPA($10^{-5}$), are utilized as the baseline algorithms. Two other state-of-the-art accelerated OPPAs, i.e., the EHPA\cite{GG_EHPA} and the APS\cite{GG_APS} are also compared against \footnote{The codes of the EHPA and the APS are available at https://github.com/caigaoub/PhysarumOptimization. We thank the authors for giving access to their source codes.}. Each algorithm will be evaluated 15 times for each graph.

\begin{figure*}
	\centering
	\subfloat[`Da-Com-1' instance .]{\includegraphics[width=1.5in]{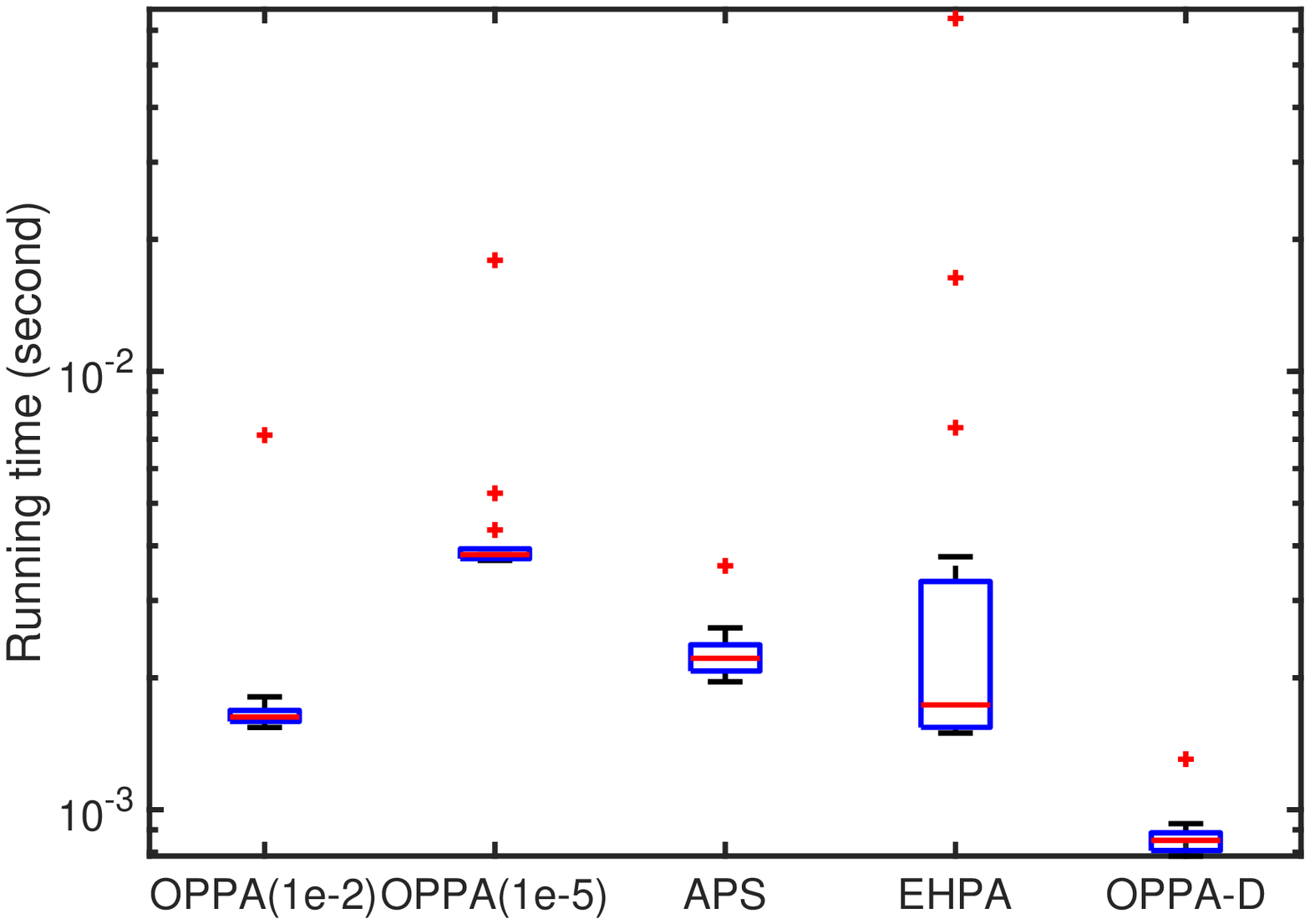}}\subfloat[`Da-Com-2' instance .]{\includegraphics[width=1.5in]{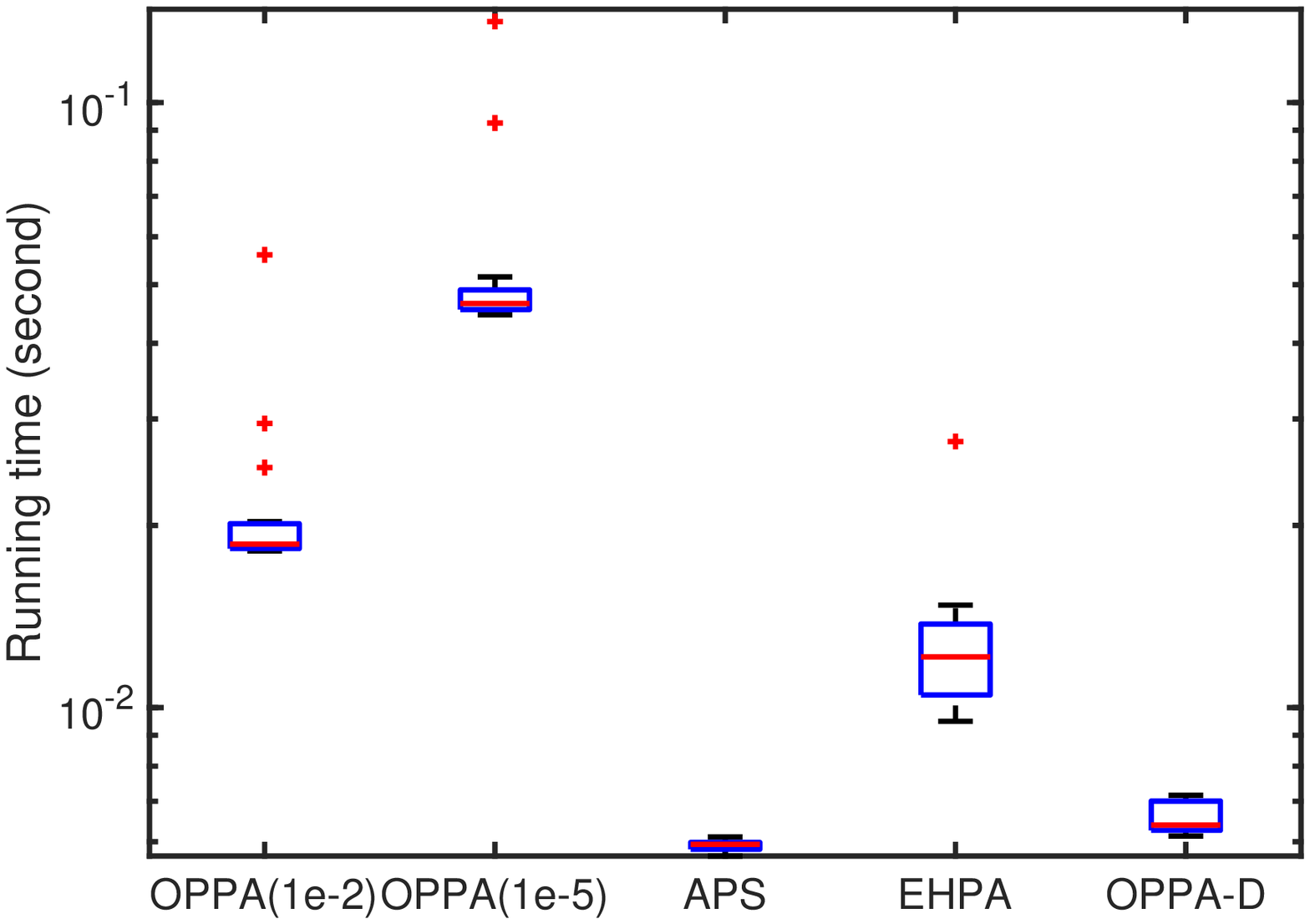}}
	\subfloat[`Da-Com-3' instance .]{\includegraphics[width=1.5in]{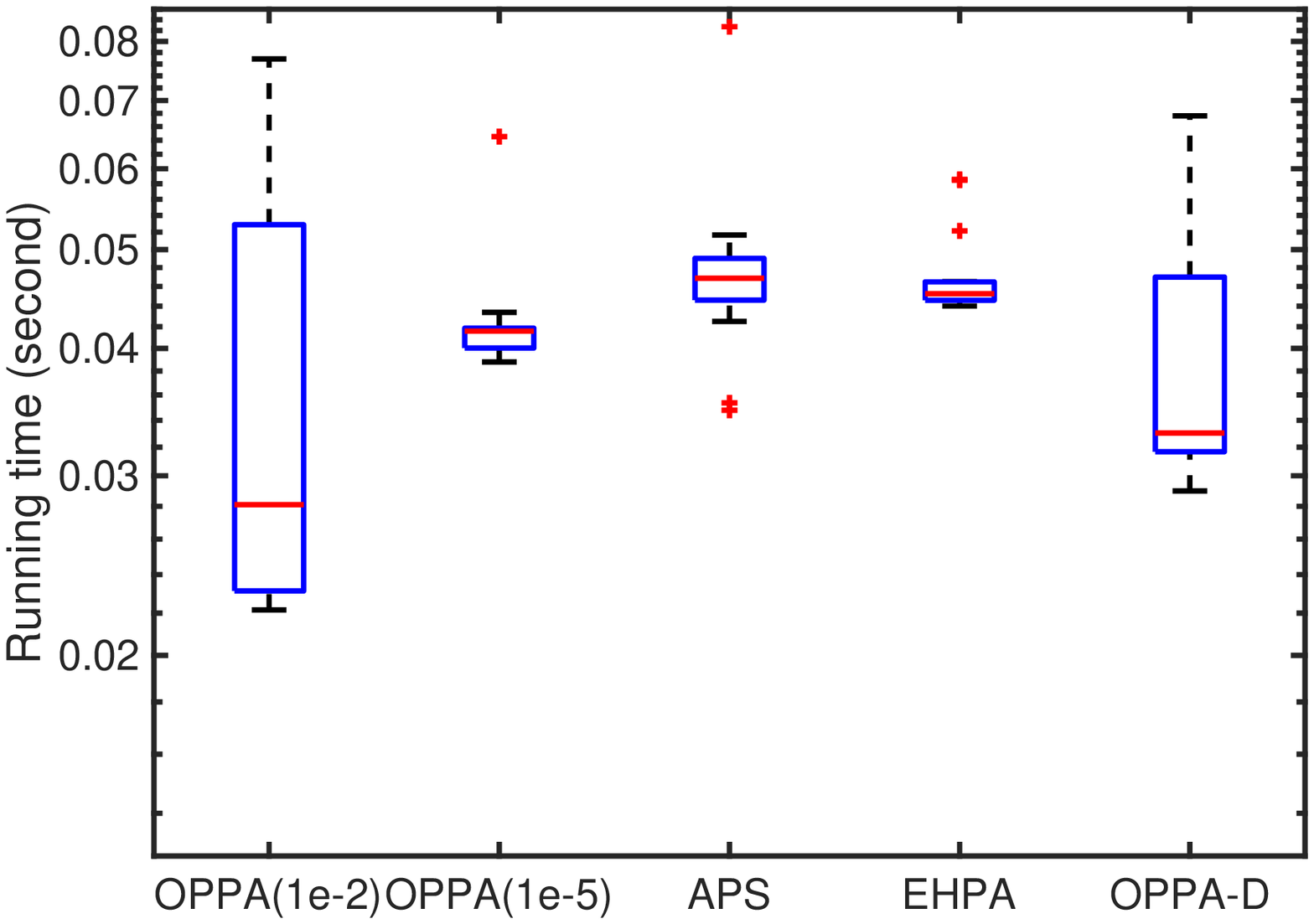}}\\
	\subfloat[`Da-Com-4' instance .]{\includegraphics[width=1.5in]{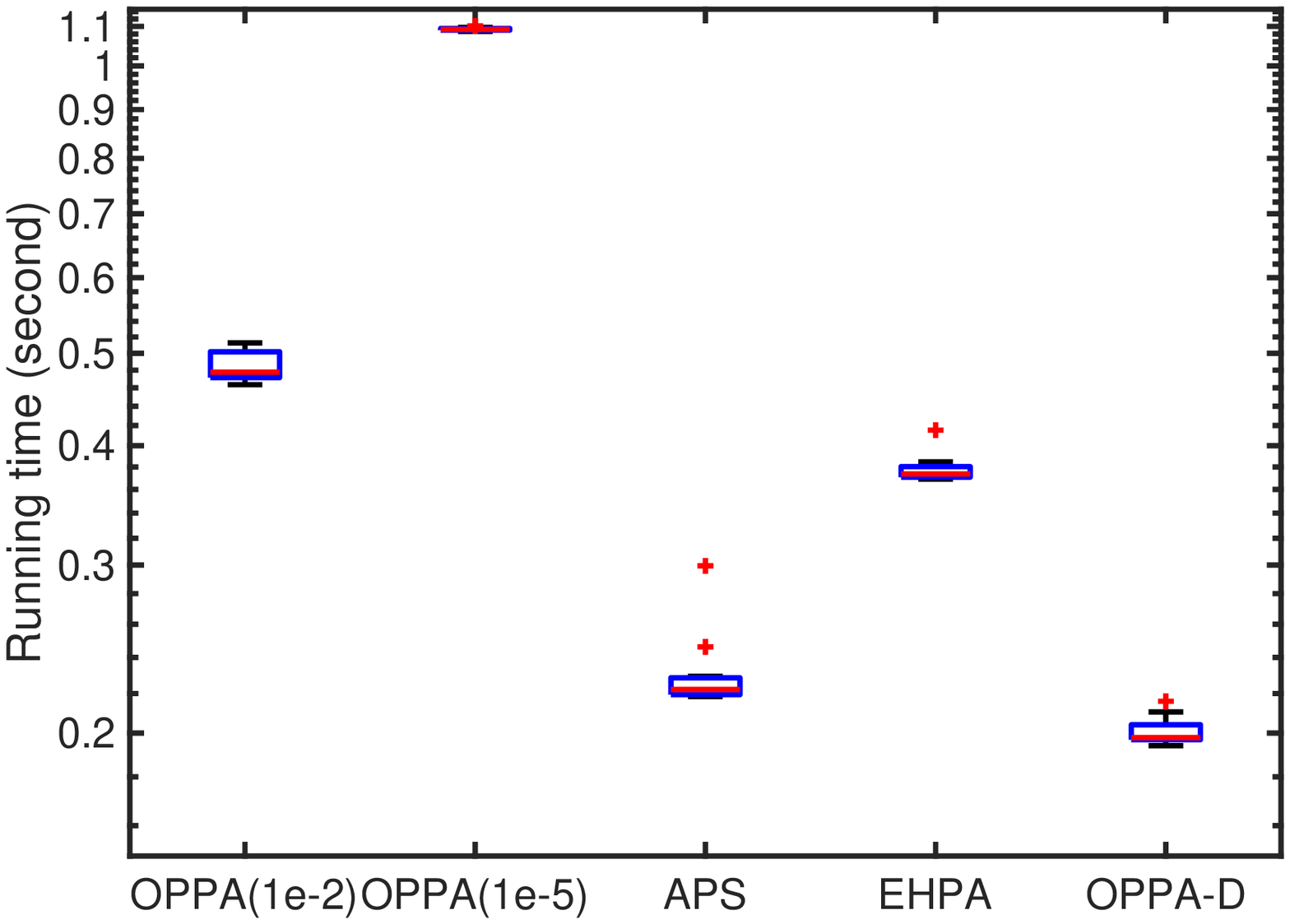}}
	\subfloat[`Da-Com-5' instance .]{\includegraphics[width=1.5in]{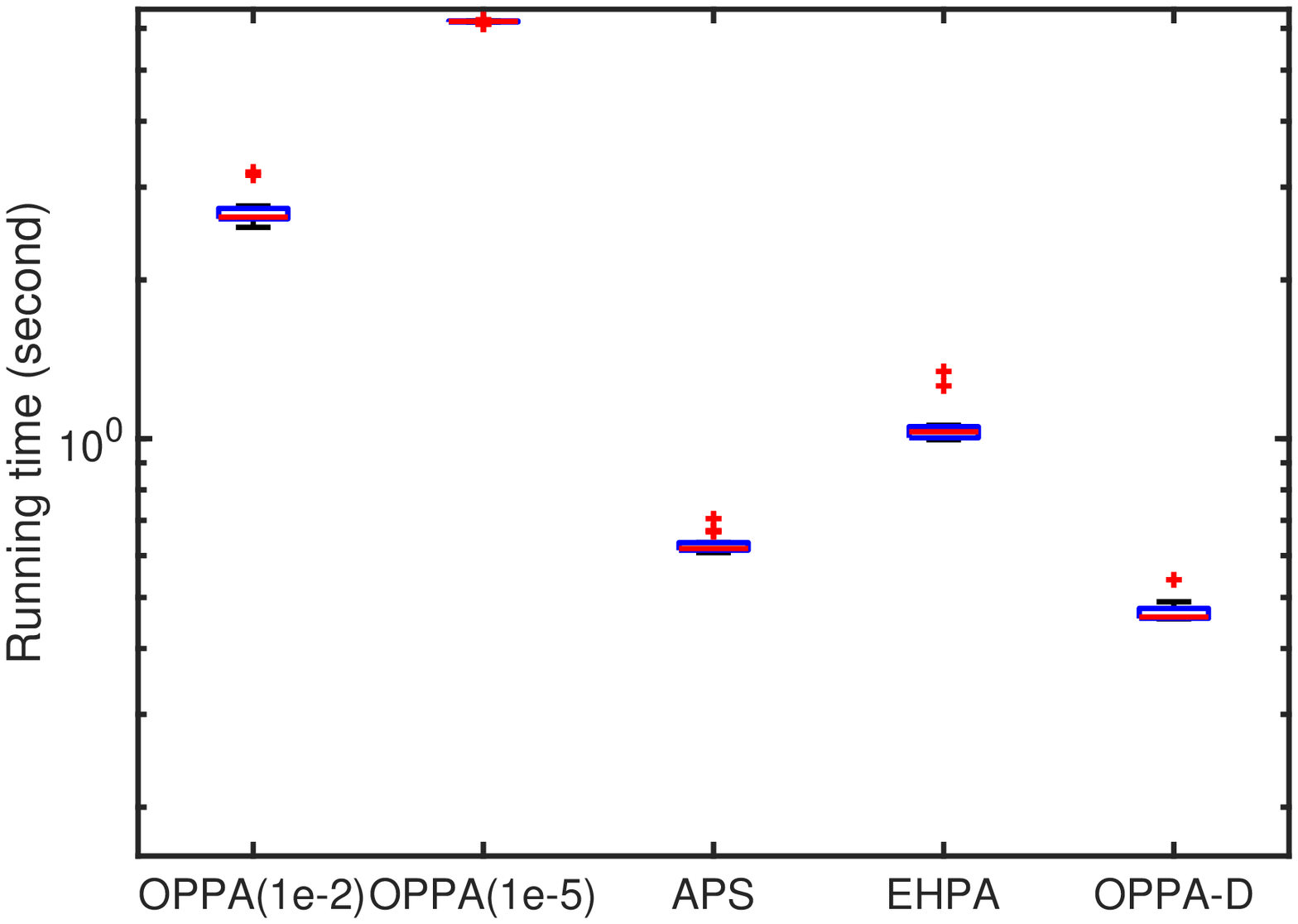}}\subfloat[`Da-Com-6' instance .]{\includegraphics[width=1.5in]{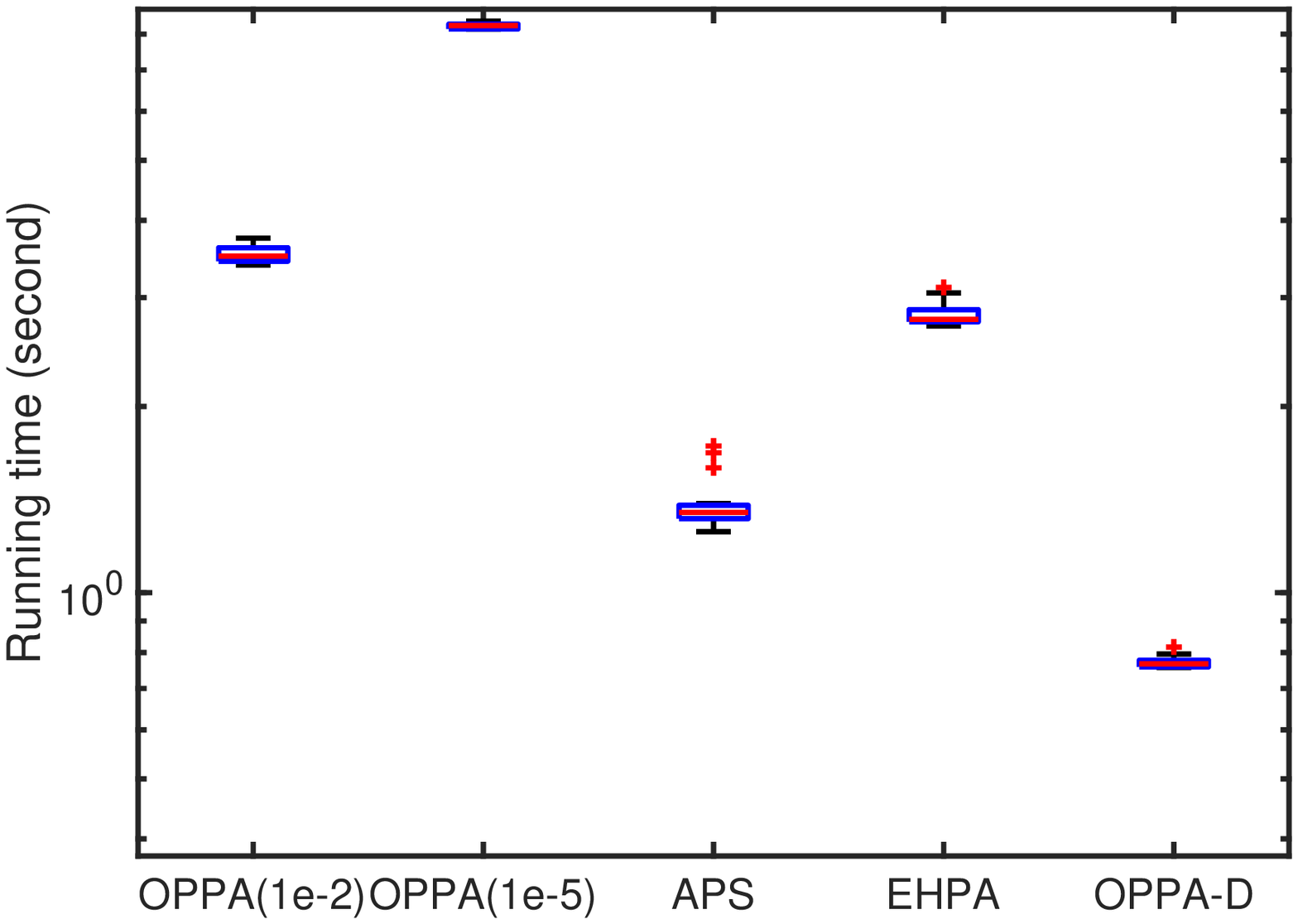}}\\
	\subfloat[`Da-Com-7' instance .]{\includegraphics[width=1.5in]{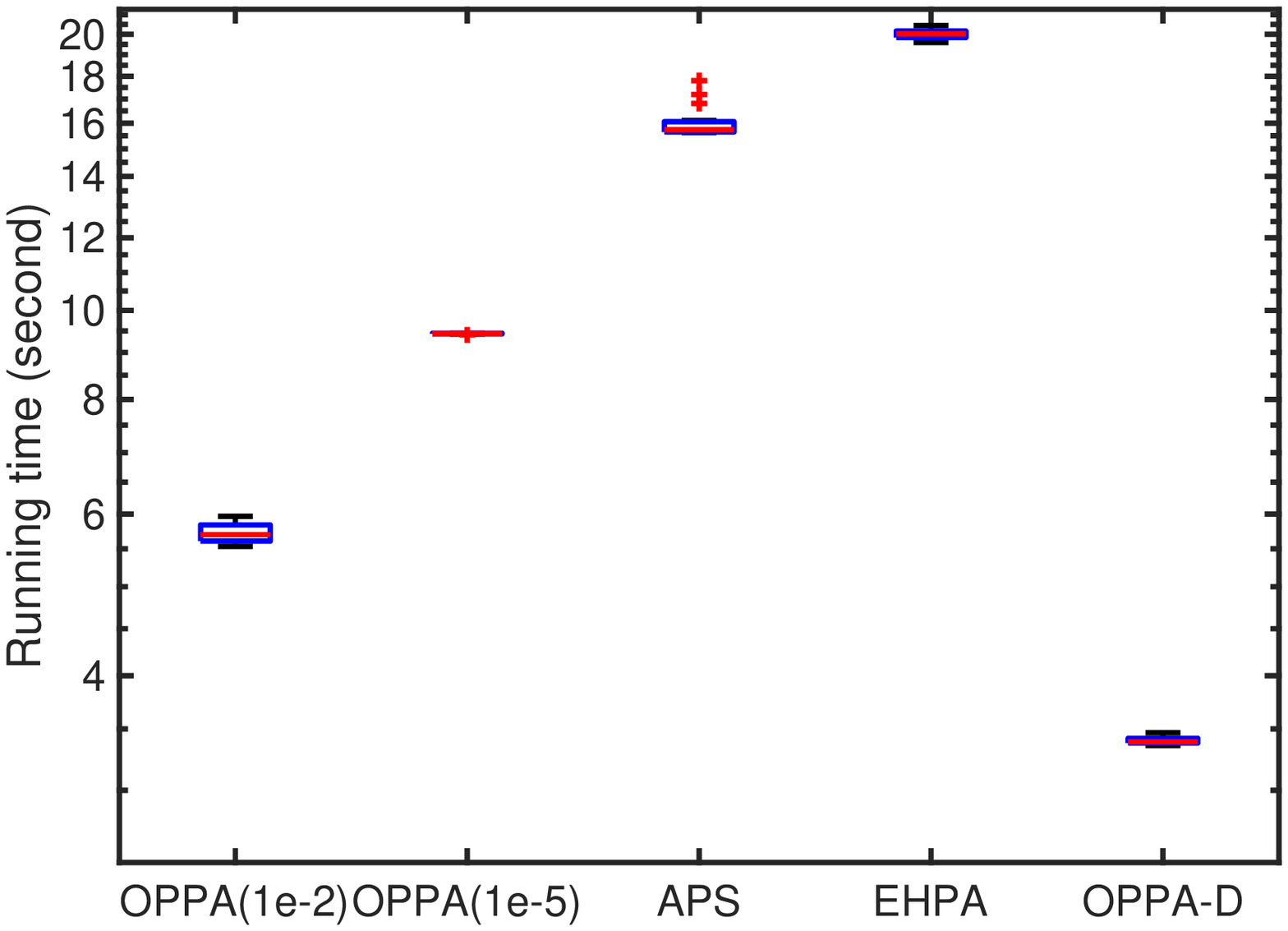}}\subfloat[`Da-Com-8' instance .]{\includegraphics[width=1.5in]{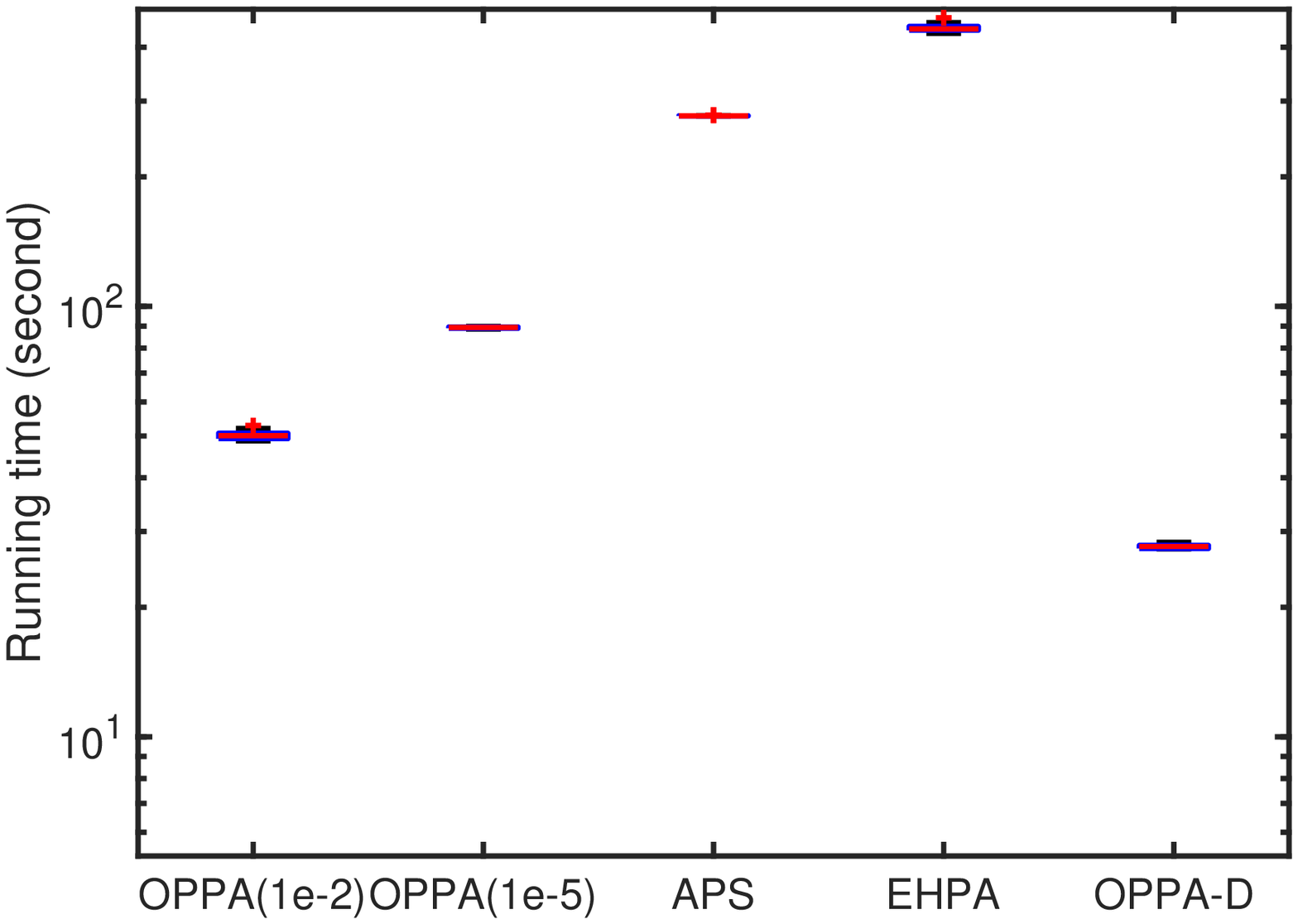}}
	\caption{Experimental results in complete graphs.}
	\label{fig.results_Da-Co}
\end{figure*}

\begin{figure*}
	\centering
	\subfloat[`Da-Sw-1' instance .]{\includegraphics[width=1.5in]{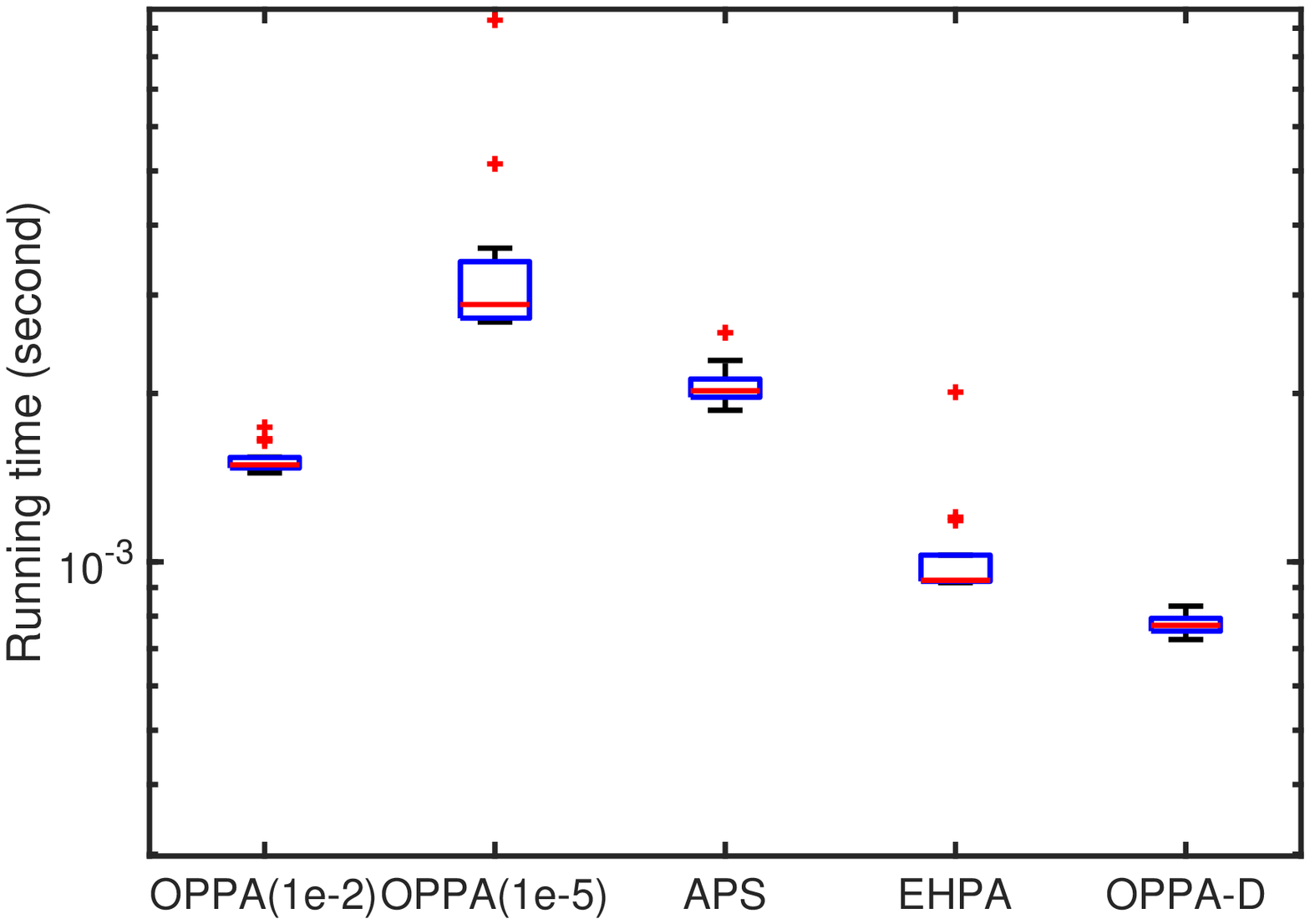}}\subfloat[`Da-Sw-2' instance .]{\includegraphics[width=1.5in]{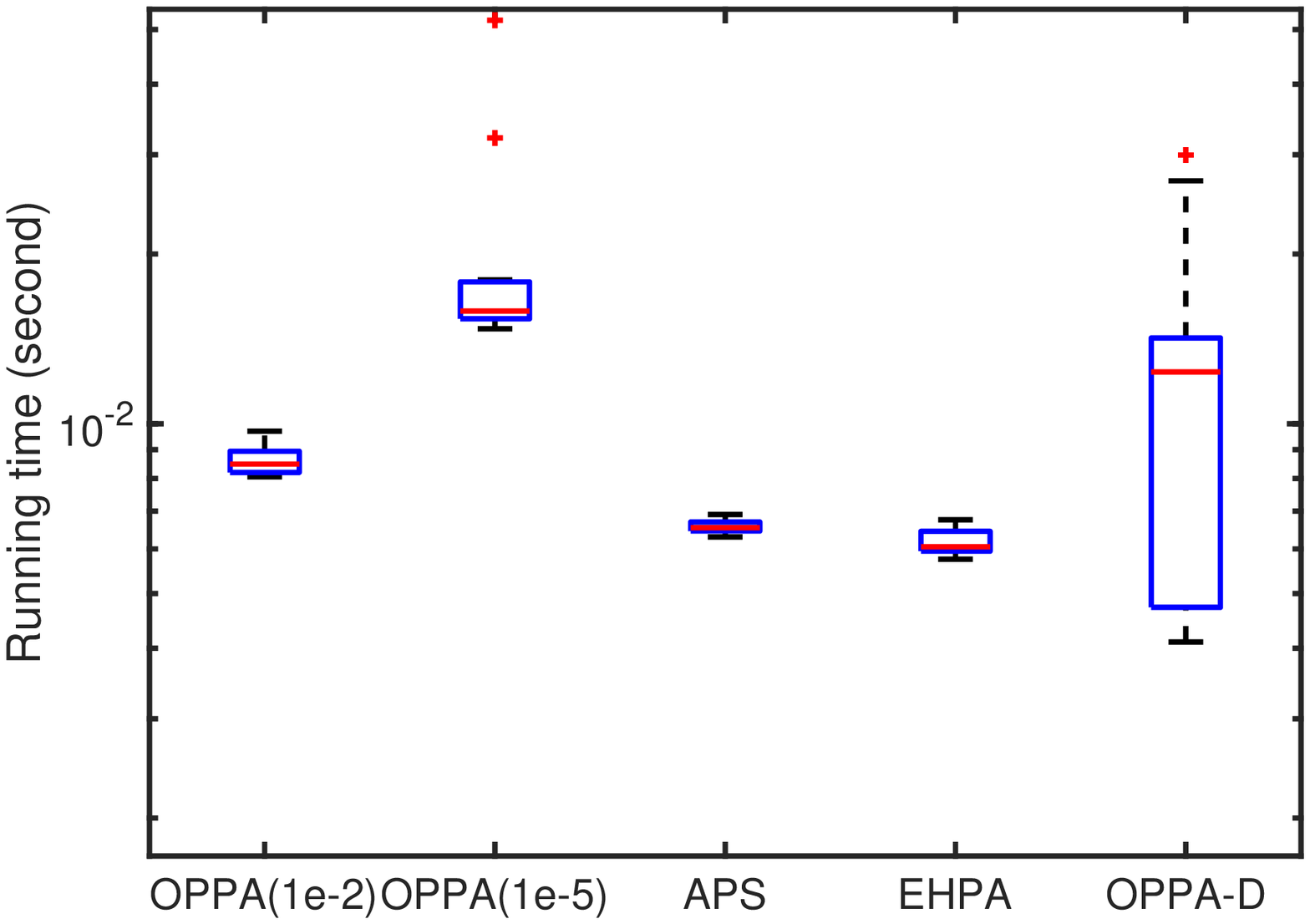}}
	\subfloat[`Da-Sw-3' instance .]{\includegraphics[width=1.5in]{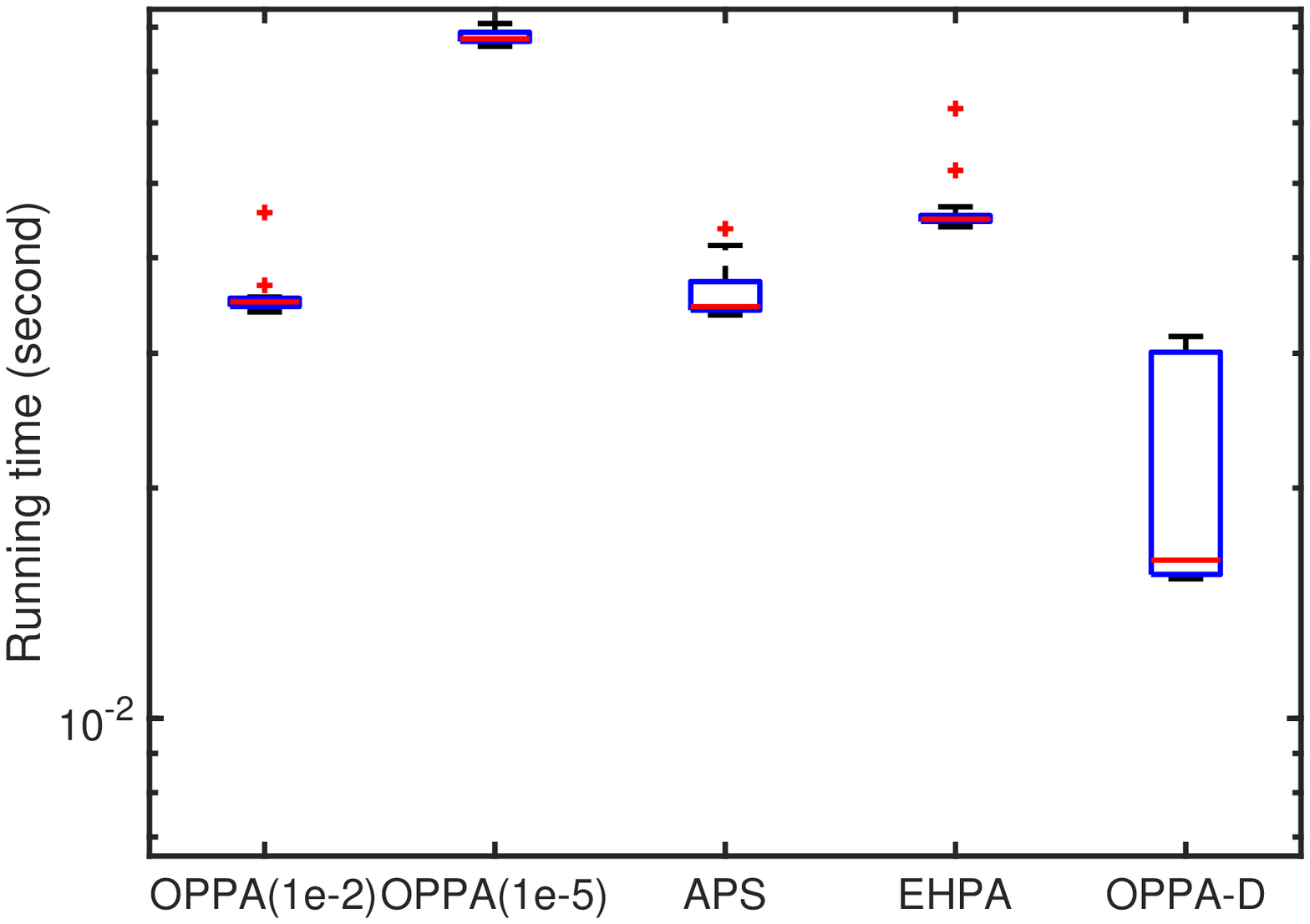}}\\
	\subfloat[`Da-Com-4' instance .]{\includegraphics[width=1.5in]{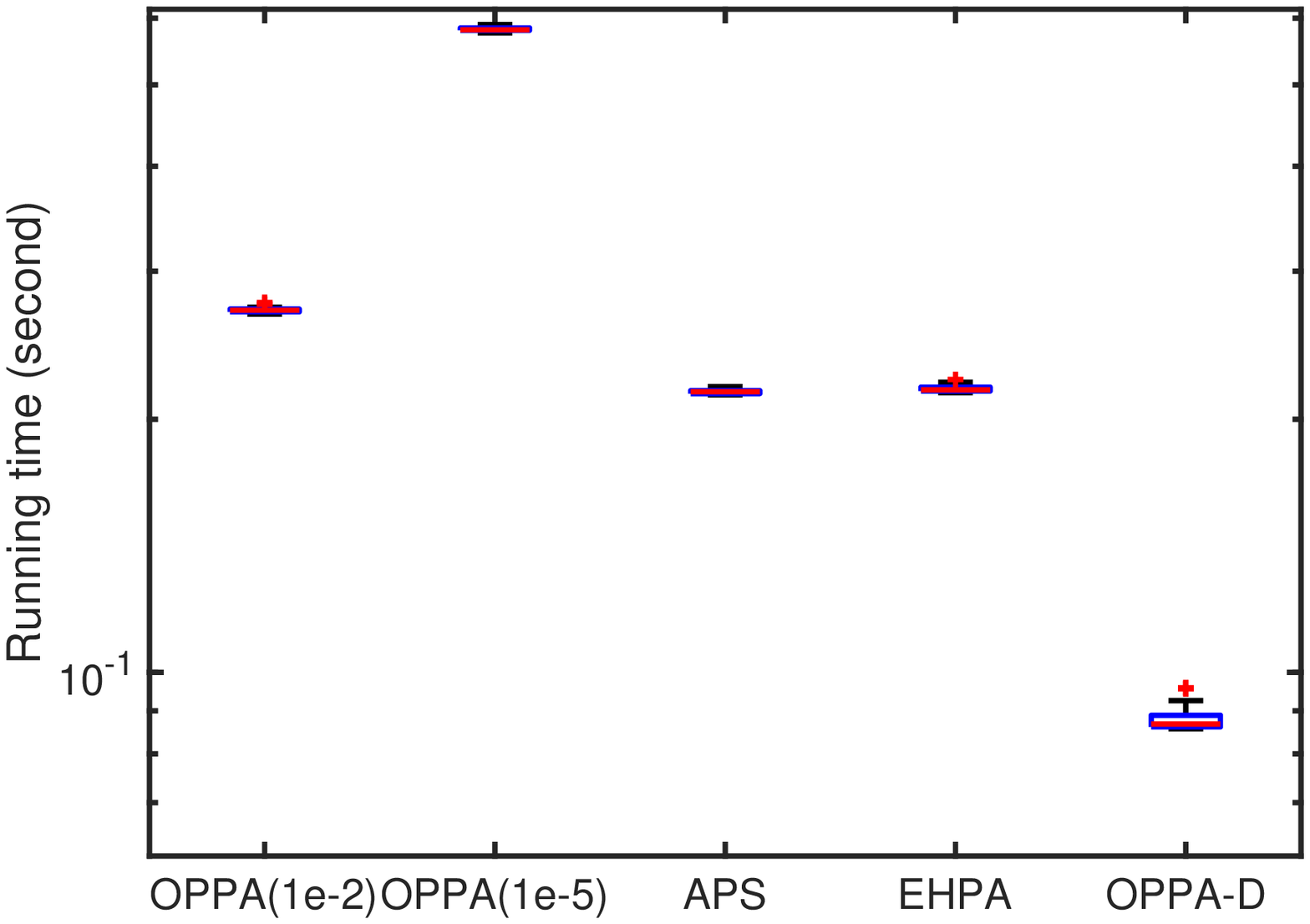}}
	\subfloat[`Da-Sw-5' instance .]{\includegraphics[width=1.5in]{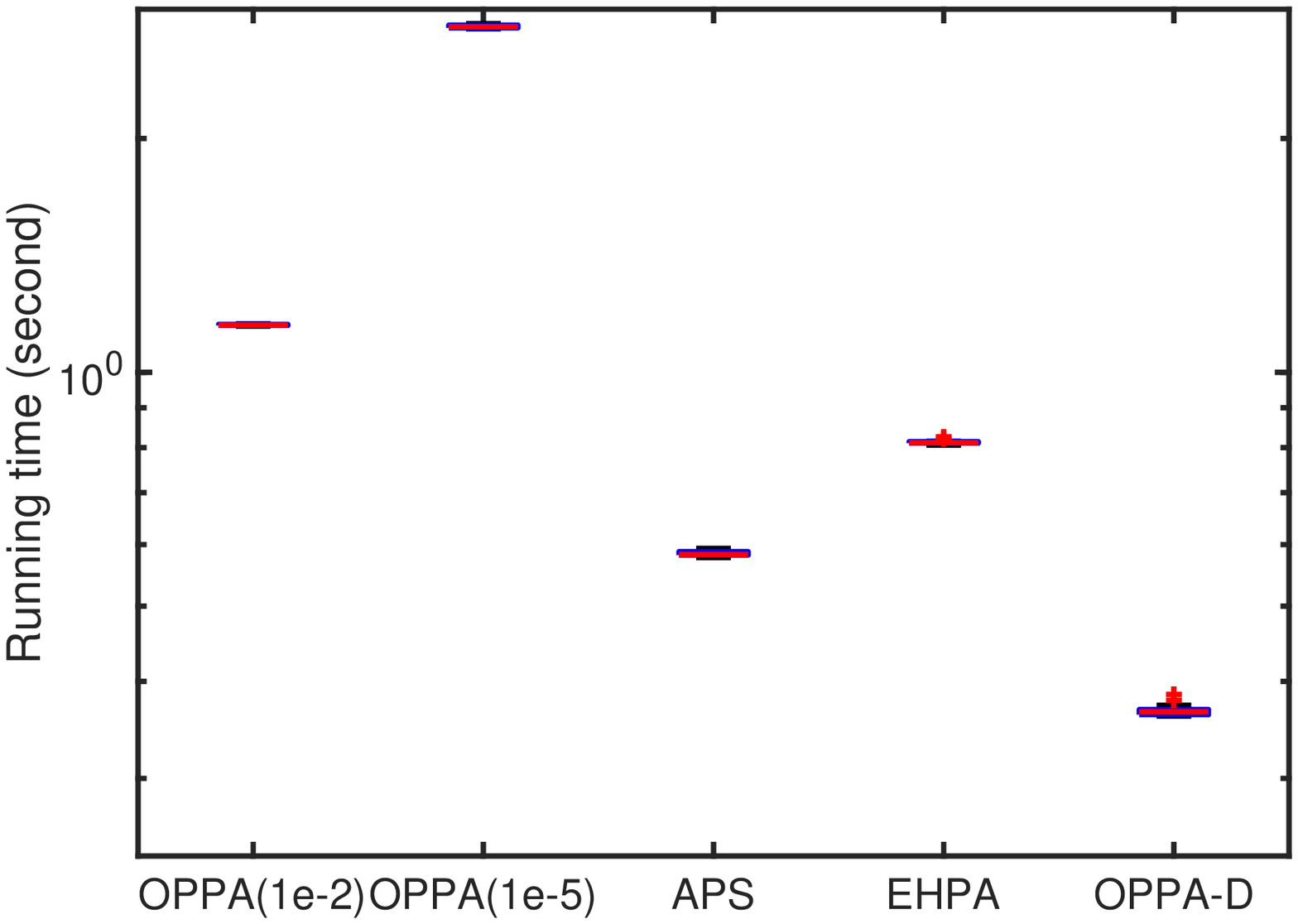}}
	\subfloat[`Da-Sw-6' instance .]{\includegraphics[width=1.5in]{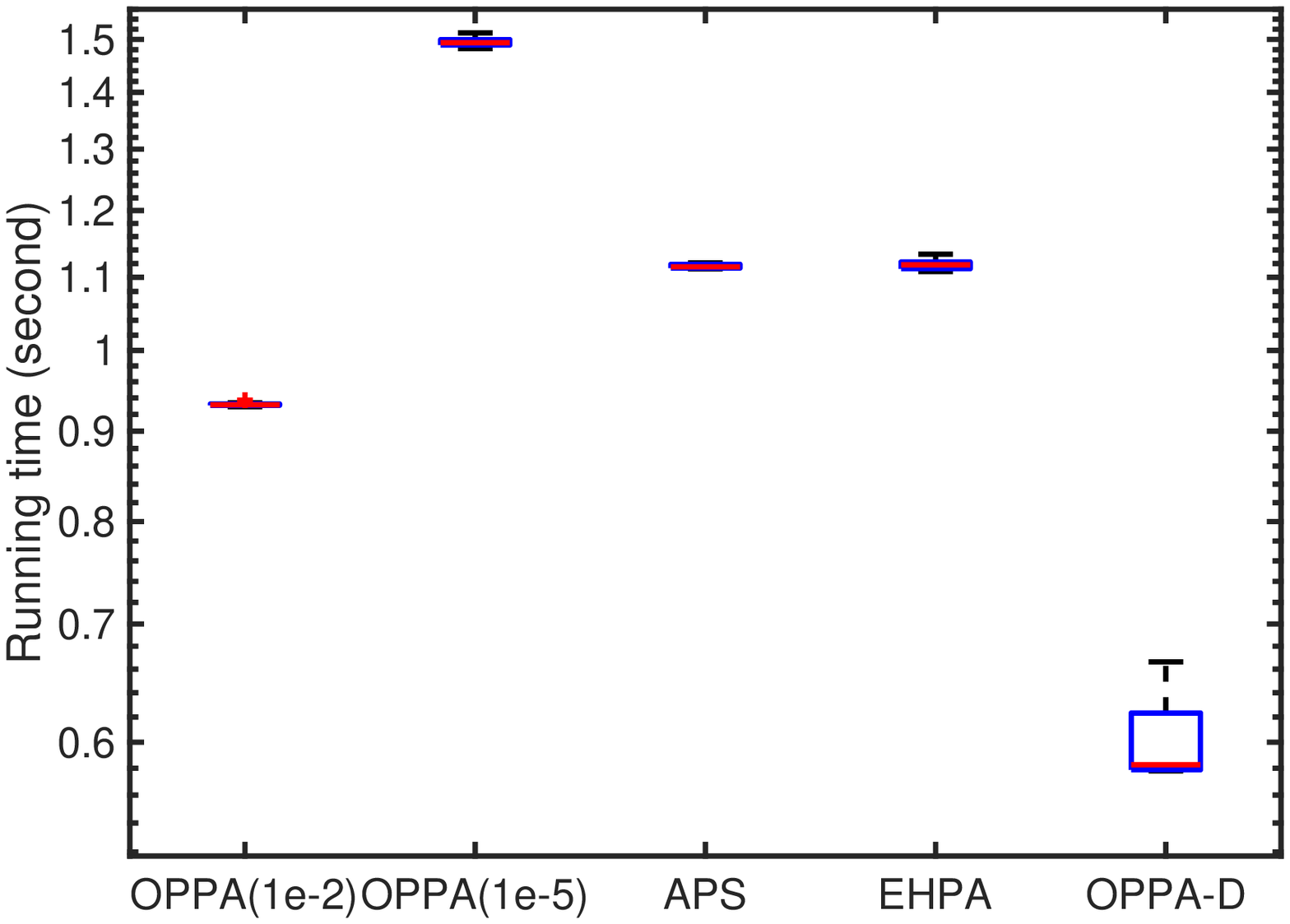}}\\
	\subfloat[`Da-Sw-7' instance .]{\includegraphics[width=1.5in]{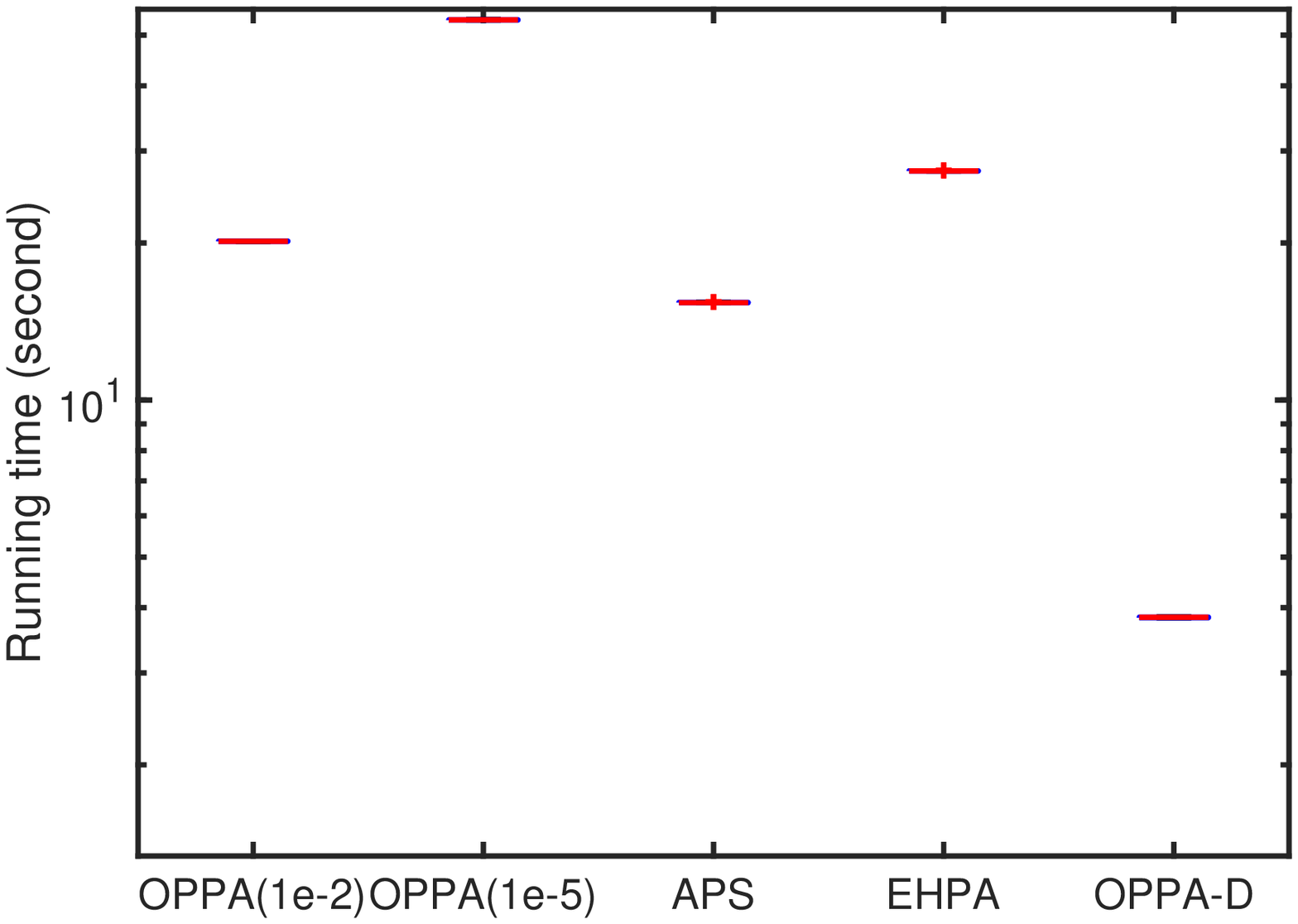}}\subfloat[`Da-Sw-8' instance .]{\includegraphics[width=1.5in]{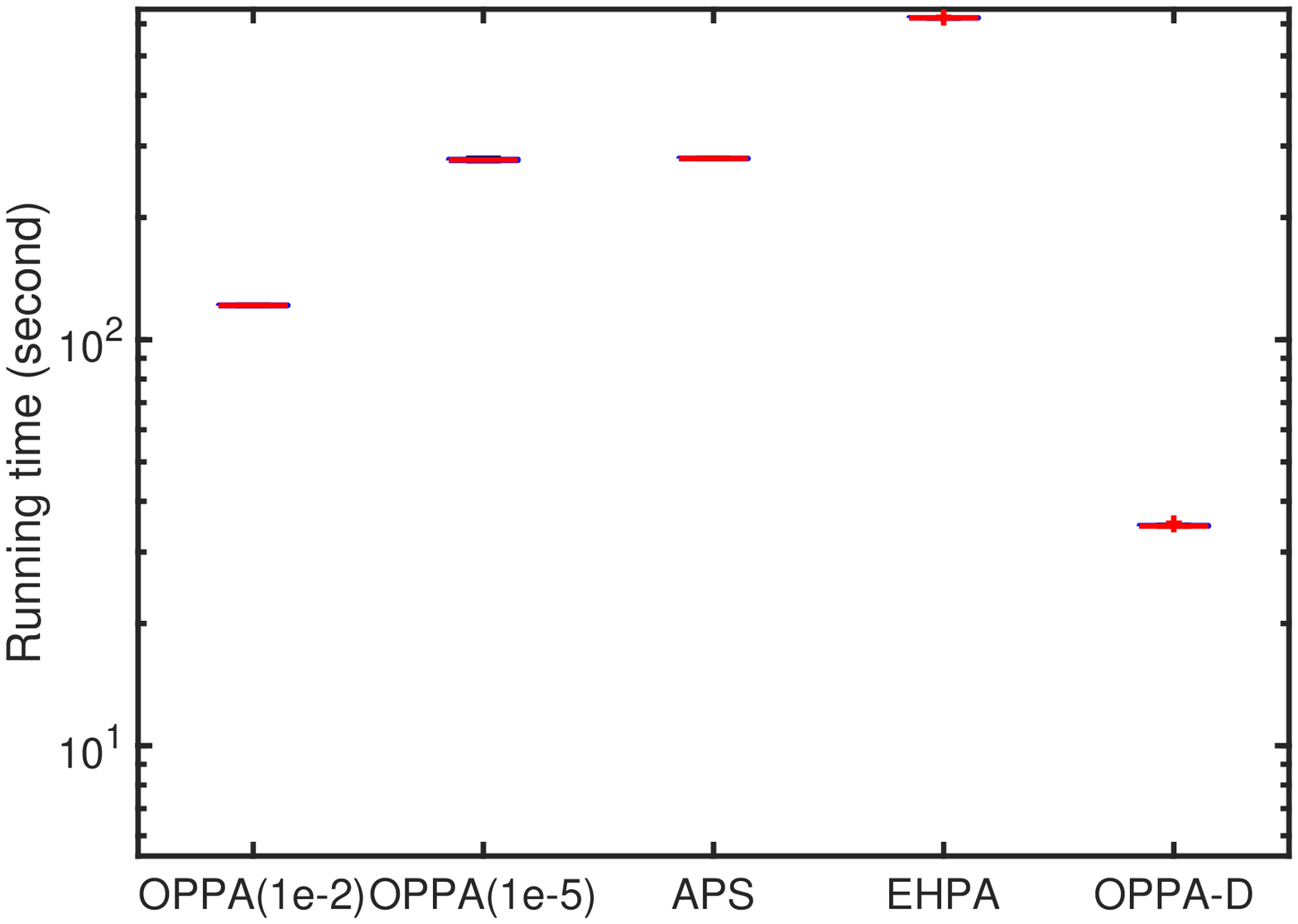}}
	\caption{Experimental results in small-world networks.}
	\label{fig.results_Da-SW}
\end{figure*}

\begin{figure*}
	\centering
	\subfloat[`Da-R-1' instance .]{\includegraphics[width=2in]{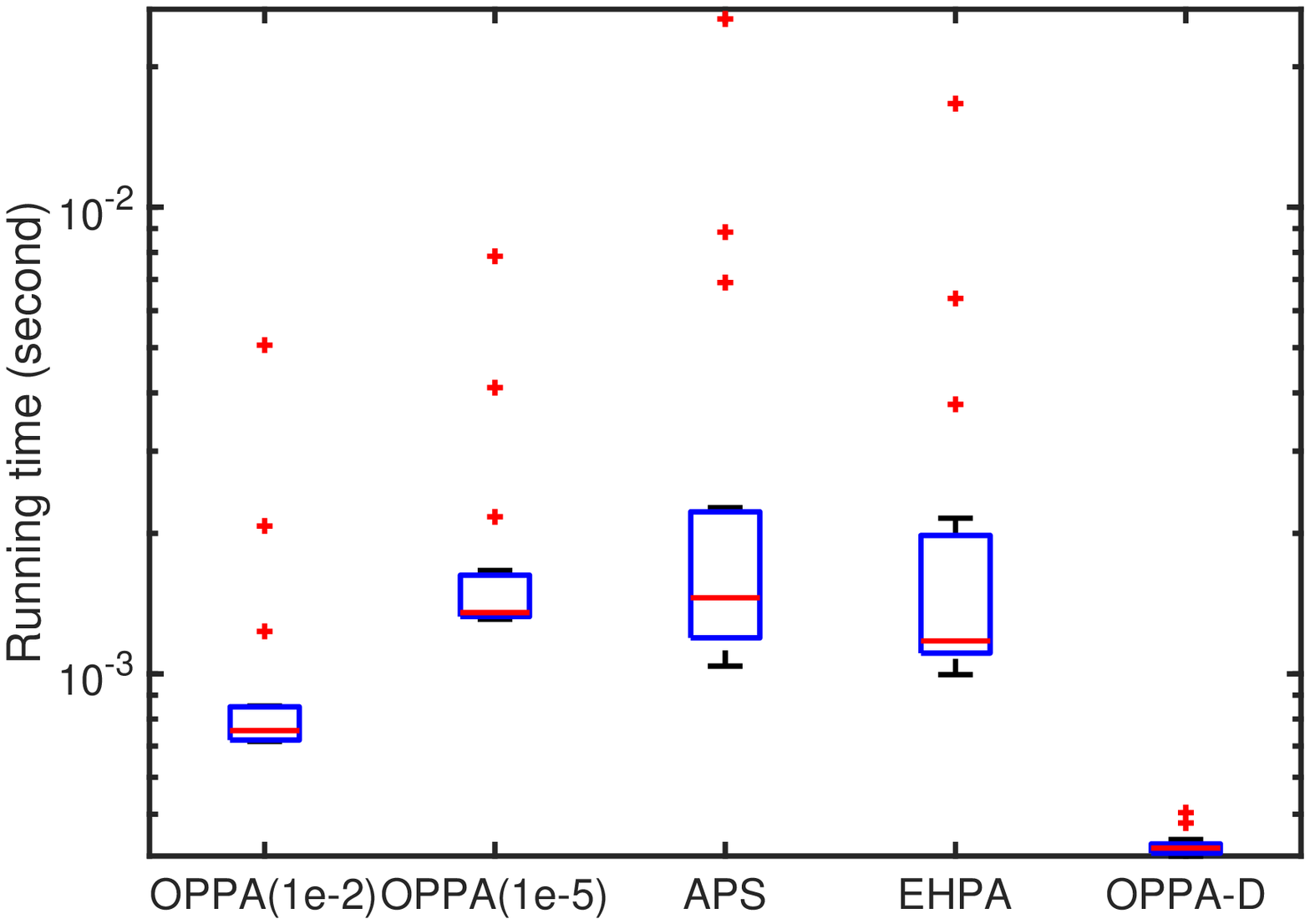}}\subfloat[`Da-R-2' instance .]{\includegraphics[width=2in]{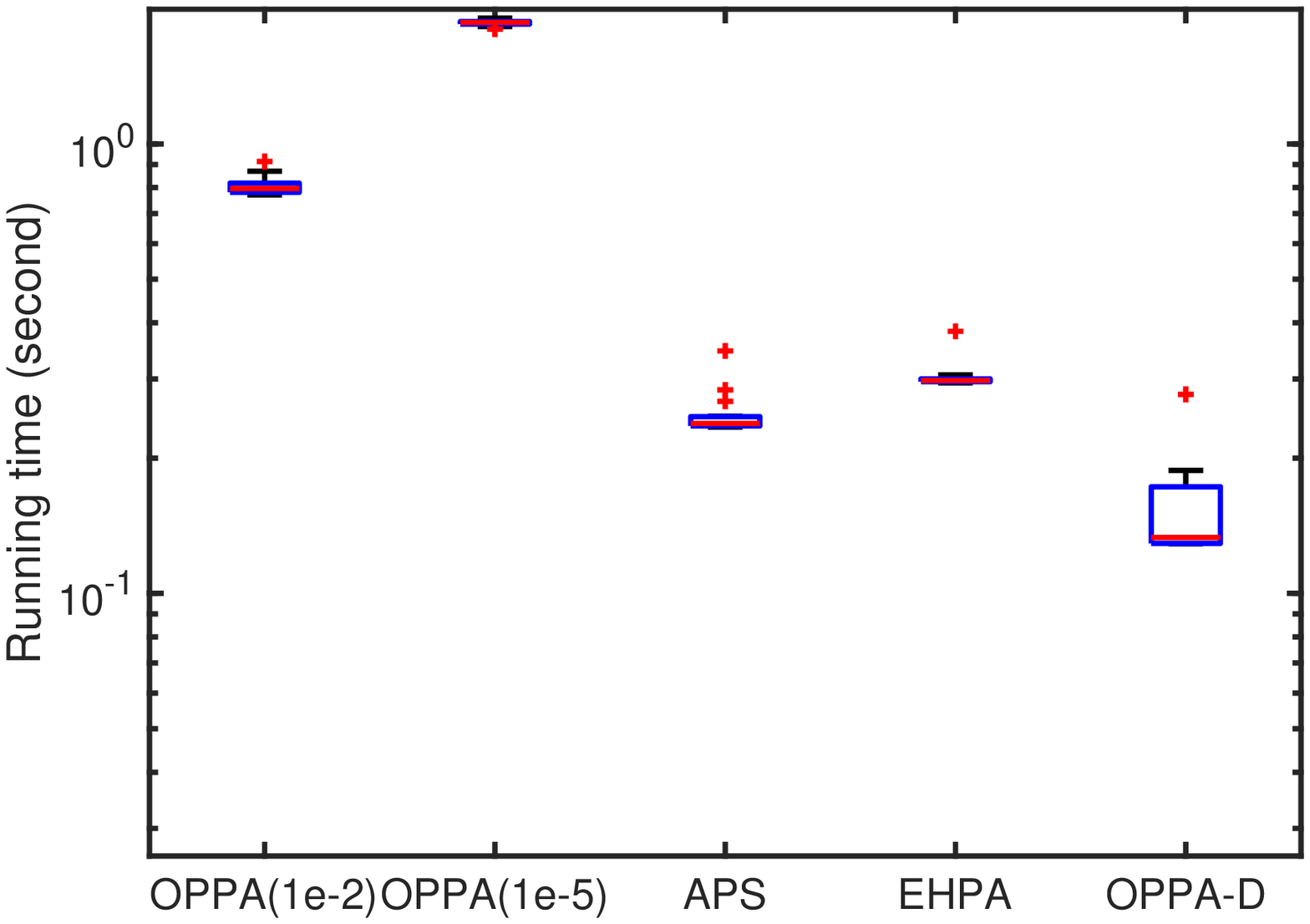}}
	\caption{Experimental results in real-world networks.}
	\label{fig.results_Da-R}
\end{figure*}

\begin{table}
	\centering
	\caption{Average running time of tested algorithms in Data Set 1}
	\resizebox{0.5\textwidth}{25mm}{
		\begin{tabular}{cccccc}
			\toprule[0.75pt]
			\multirow{2}{*}{Instacnes} 	& \multicolumn{5}{c}{Running time (Second)}\\					
			&	OPPA($10^{-2}$)&	OPPA($10^{-5}$)&	EHPA&	APS&	OPPA-D(Ours)	\\\hline
			Da-Com-1	&	0.0020&	0.0049&	0.0073&	0.0023&	\textbf{0.0009}	\\
			Da-Com-2	&	0.0223&	0.0557&	0.0131&	\textbf{0.005}9&	0.0066	\\
			Da-Com-3	&	0.0400&	0.0425&	0.0474&	0.0478&	\textbf{0.0398}	\\
			Da-Com-4	&	0.4839&	1.0926&	0.3771&	0.2293&	\textbf{0.2006}	\\
			Da-Com-5	&	2.7072&	6.1812&	1.0617&	0.6322&	\textbf{0.4698}	\\
			Da-Com-6	&	3.5256&	8.2404&	2.8248&	1.3938&	\textbf{0.7721}	\\
			Da-Com-7	&	5.7222&	9.4358&	20.0010&	16.0612&	\textbf{3.4003}	\\
			Da-Com-8	&	50.1900&	89.3060&	444.2410&	277.3507&	\textbf{27.6500}	\\
			Da-SW-1	&	0.0015&	0.0035&	0.0010&	0.0021&	\textbf{0.0008}	\\
			Da-SW-2	&	0.0086&	0.0195&	\textbf{0.0062}&	0.0066&	0.0125	\\
			Da-SW-3	&	0.0357&	0.0777&	0.0466&	0.0362&	\textbf{0.0217}	\\
			Da-SW-4	&	0.2700&	0.5818&	0.2176&	0.2157&	\textbf{0.0883}	\\
			Da-SW-5	&	1.1505&	2.7874&	0.8128&	0.5841&	\textbf{0.3672}	\\
			Da-SW-6	&	0.9320&	1.4951&	1.1180&	1.1162&	\textbf{0.5987}	\\
			Da-SW-7	&	20.1512&	53.4956&	27.4835&	15.3748&	\textbf{3.8334}	\\
			Da-SW-8	&	121.5098&	277.3977&	620.7715&	279.5441&	\textbf{34.8306}	\\
			Da-R-1	&	0.0012&	0.0021&	0.0028&	0.0039&	\textbf{0.0004}	\\
			Da-R-2	&	0.8085&	1.8604&	0.3038&	0.2507&	\textbf{0.1533}	\\
			\bottomrule[0.75pt] 
		\end{tabular}
	}
	\\\footnotesize{$^a$Values presented in the table are the average value of 15 trials.}
	\label{table.exp1_time}
\end{table}

\begin{table}
	\centering
	\caption{Average number of iterations of tested algorithms in Data Set 1}
	\resizebox{0.5\textwidth}{25mm}{
		\begin{tabular}{cccccc}
			\toprule[0.75pt]
			\multirow{2}{*}{Instacnes} 	& \multicolumn{5}{c}{Number of iterations}\\					
			&	OPPA($10^{-2}$)&	OPPA($10^{-5}$)&	EHPA&	APS&	OPPA-D(Ours)	\\\hline
			Da-Com-1	&	26&	58&	12&	16&	12	\\
			Da-Com-2	&	57&	134&	21&	16&	19	\\
			Da-Com-3	&	27&	49&	16&	16&	18	\\
			Da-Com-4	&	58&	139&	23&	17&	25	\\
			Da-Com-5	&	134&	337&	26&	26&	25	\\
			Da-Com-6	&	96&	238&	33&	20&	22	\\
			Da-Com-7	&	31&	55&	19&	16&	19	\\
			Da-Com-8	&	41&	75&	26&	17&	23	\\
			Da-SW-1	&	24&	47&	8&	16&	11	\\
			Da-SW-2	&	27&	50&	14&	16&	11	\\
			Da-SW-3	&	41&	89&	20&	16&	17	\\
			Da-SW-4	&	34&	73&	16&	16&	11	\\
			Da-SW-5	&	62&	149&	25&	18&	20	\\
			Da-SW-6	&	27&	43&	16&	16&	17	\\
			Da-SW-7	&	116&	305&	29&	16&	22	\\
			Da-SW-8	&	105&	233&	38&	36&	30	\\
			Da-R-1	&	27&	53&	9&	16&	11	\\
			Da-R-2	&	133&	330&	37&	44&	24	\\		
			\bottomrule[0.75pt] 
		\end{tabular}
	}
	\\\footnotesize{$^a$Values presented in the table are the average value of 15 trials.}
	\label{table.exp1_ite}
\end{table}

All the evaluated algorithms successfully solved for the shortest path in all 15 instances. The running time of them are presented in Fig.\ref{fig.results_Da-Co}, \ref{fig.results_Da-SW}, \ref{fig.results_Da-R}. Tables \ref{table.exp1_time} and \ref{table.exp1_ite} summarize the running time and the number of iterations, respectively. 

In general, the OPPA-D is superior to the baseline algorithms in terms of efficiency. In the complete graphs, as recorded in Fig.\ref{fig.results_Da-Co} and Table \ref{table.exp1_time}, the OPPA-D outperforms all other baseline algorithms except for `Da-Com-2' where the APS is slightly faster than the OPPA-D. In the other instances, the OPPA-D is 56.40\%, 0.37\%, 12.48\%, 25.69\%, 44.61\%, 40.58\%, and 44.91\% faster than the next fastest algorithm, respectively. In the small-world graphs, according to Fig.\ref{fig.results_Da-SW} and Table \ref{table.exp1_time}, the OPPA-D is more efficient than all the baseline algorithms except for `Da-Sw-2' where it ranks fourth. In other instances, the OPPA-D is 26.21\%, 39.19\%, 59.06\%, 37.14\%, 35.76\%, 75.07\%, and 71.34\% faster than the second-ranked algorithm, respectively. In the real-world network, according to Fig.\ref{fig.results_Da-R} and Table \ref{table.exp1_time}, the OPPA-D is the fastest algorithm. It is 63.09\% and 38.84\% more efficient than the algorithm in second place, respectively. 

Though the accelerated OPPAs, i.e., the EHPA and the APS, are faster than the OPPA($10^{-2}$) and OPPA($10^{-5}$) in most instances, they do not out-perform the OPPA($\epsilon=10^{-5}$) in the 2000-nodes and 5000-node complete graphs and 5000-node small-world network. One interesting finding in this experiment is that, according to Tables \ref{table.exp1_time} and \ref{table.exp1_ite}, while the number of iterations is small it does not necessarily translate to a shorter running time. For example, the OPPA($10^{-2}$) takes 41 iterations to converge while the APS only take 17 iterations for `Da-Com-8'. However, the OPPA($10^{-2}$) is approximately five times faster than the APS. Therefore, when comparing different accelerated OPPAs, the running time, instead of the number of iterations, would be a more appropriate evaluation metric. 

\subsection{Comparing different termination criteria}
\label{sec.4.2}
In the above experiments, we observe that the running time of the OPPA is highly sensitive to the current stopping criteria, i.e., $\sum_{i}\sum_{j\neq i}D_{ij}\leq\epsilon$. According to Table \ref{table.exp1_time}, the OPPA($10^{-2}$) is more efficient than the OPPA($10^{-5}$); both algorithms solve for the shortest-path in all the instances successfully. Recall that the OPPA-D is fundamentally the OPPA with a newly proposed termination criterion. Thus, to validate the superiority of the proposed criterion, i.e. executing the OPPA until the $L_{D-Path}$ reamins steady for $K$ iteration, we conduct further experiments.

\noindent\textbf{The evaluated criteria:} 1) The traditional termination criterion with $\epsilon=10^{-1}, 10^{-2}, \cdots, 10^{-5}$ are tested. 2) The proposed termination criterion with $K=5, 10, 15, 20, 30$ are also tested.

\noindent\textbf{Data Set 2:} We test the termination criteria in randomly generated complete graphs. The size of the graphs are 10, 100, and 500. For each size, 50 graphs are randomly generated with the length of the edge varying from 1 to 10000.

\begin{table*}
	\centering
	\caption{Experimental results in Data Set 2--Success rate.}
	\resizebox{0.9\textwidth}{9.5mm}{
		\begin{tabular}{ccccccccccc}
			\toprule[0.75pt]
			\multirow{2}{*}{Size} & \multicolumn{5}{c}{The traditional criteria ($\epsilon$)} & \multicolumn{5}{c}{The proposed criteria ($K$)} \\\cline{2-6}\cline{7-11}
			&	($10^{-1}$)	&	($10^{-2}$)	&	($10^{-3}$)	&	($10^{-4}$)	&	\multicolumn{1}{c|}{($10^{-5}$)}&	5	&	10	&	15	&	20	&	30	\\\hline
			10	&	96\%(2$|$0)	&	100\%(0$|$0)	&	100\%(0$|$0)	&	100\%(0$|$0)	&	100\%(0$|$0)	&	96\%(2$|$0)	&	96\%(2$|$0)	&	100\%(0$|$0)	&	100\%(0$|$0)	&	100\%(0$|$0)	\\
			100	&	98\%(1$|$0)	&	100\%(0$|$0)	&	96\%(0$|$2)	&	92\%(0$|$4)	&	88\%(0$|$6)	&	60\%(20$|$0)	&	88\%(6$|$0)	&	90\%(5$|$0)	&	96\%(2$|$0)	&	100\%(0$|$0)	\\
			500	&	100\%(0$|$0)	&	100\%(0$|$0)	&	98\%(0$|$1)	&	96\%(0$|$2)	&	96\%(0$|$2)	&	64\%(18$|$0)	&	92\%(4$|$0)	&	98\%(1$|$0)	&	100\%(0$|$0)	&	100\%(0$|$0)	\\
			\bottomrule[0.75pt] 
		\end{tabular}
	}
	\\\footnotesize{$^a$ The values in the table have a format `Value1\%(Value2$|$Value3)', where Value1 is the success rate out of 50 cases, Value2 represents the number of the cases that the algorithm does not successfully solve, Value3 is the number of cases where the algorithm could not reach the termination criteria though sufficient computational time.}
	\label{table.exp2-srate}
\end{table*}

\begin{table*}
	\centering
	\caption{Experimental results in Data Set 2--Running time.}
	\resizebox{0.75\textwidth}{7mm}{
		\begin{tabular}{ccccccccccc}
			\toprule[0.75pt]
			\multirow{2}{*}{Size} & \multicolumn{5}{c}{The traditional criteria ($\epsilon$)} & \multicolumn{5}{c}{The proposed criteria ($K$)} \\\cline{2-6}\cline{7-11}
			&	($10^{-1}$)	&	($10^{-2}$)	&	($10^{-3}$)	&	($10^{-4}$)	&	\multicolumn{1}{c|}{($10^{-5}$)}	&	5	&	10	&	15	&	20	&	30	\\\hline
			10	&	0.0003 	&	0.0013 	&	0.0045 	&	0.0078 	&	0.0111 	&	0.0002 	&	0.0003 	&	0.0003 	&	0.0004 	&	0.0006 	\\
			100	&	0.0130 	&	0.0410 	&	0.0913 	&	0.0978 	&	0.1061 	&	0.0043 	&	0.0065 	&	0.0084 	&	0.0105 	&	0.0141 	\\
			500	&	0.3203 	&	0.9352 	&	1.4508 	&	1.7348 	&	2.4032 	&	0.1049 	&	0.1441 	&	0.1806 	&	0.2179 	&	0.2877 	\\
			\bottomrule[0.75pt] 
		\end{tabular}
	}
	\\\footnotesize{$^a$ The values presented in the table are the average values of running time (seconds) of only the success cases from the 50 cases.}
	\label{table.exp2-runtime}
\end{table*}

Tables \ref{table.exp2-srate} and \ref{table.exp2-runtime} report the experimental results. From Table \ref{table.exp2-srate}, different settings towards parameters $\epsilon$ and $K$ may lead to different success rates. One interesting thing of note is that when the graph size is 100 or 500, as $\epsilon$ decreases, despite the decrease in the number of failed attempts, the algorithm could not reach the traditional termination criterion with a low $\epsilon$ in certain cases. However, the proposed stopping criterion does not have the above issue. 

According to Table \ref{table.exp2-runtime}, one could conclude that, from the perpective of running time, the traditional termination criterion is more sensitive to the parameter $\epsilon$, compared with the proposed stopping criterion's sensitivity to $K$. For example, when the graph size is 500, when $\epsilon$ decreases from $10^{-1}$ to $10^{-5}$, the running time increases from 0.32 seconds to 2.40 seconds (by 650\%); however, when $K$ increases from $5$ to $30$, the running time increases from 0.10 seconds to 0.29 seconds (by 190\%).

In general, in terms of of success rate, the traditional termination criterion with $\epsilon=10^{-2}$ and the proposed termination criterion with $K=30$ would be good choices, both of which achieve 100\% success rate for three different graph sizes (each of the graph size has 50 cases). After comparing our experimental results of the running time for both criteria, one would find that the proposed criterion with $K=30$ would be the better choice.

\subsection{Analyzing the improvements made by accelerated OPPAs}
\label{sec.4.3}
In the previous sub-section, we demonstrate that the proposed criterion is better than the traditional criterion. However, the running time of the OPPA for both stopping criteria is still inevitably dependent on the initial parameters. It is logical to deduce that the above findings still holds when these criteria are applied to other accelerated OPPAs. Thus, in this section, we want to compare the accelerated OPPAs with the OPPA by excluding the termination criteria.

\subsubsection{The proposed evaluation method}
\label{sec.4.3.1}
Using the defined concepts of the D-Path and T-Point, we propose our evaluation method:
\begin{enumerate}[i.]
	\item When generating the data set, calculate the shortest path of each graph.
	\item Evaluate the algorithm in the data set. In each iteration of the algorithm, calculate the $L_{D-Path}$. If the $L_{D-Path}$ of this iteration is equal to the shortest path length, record the corresponding running time and proceed to Step iii; otherwise, repeat Step ii.
	\item Execute the algorithm for 50 or more iterations to ensure the convergence of the algorithm. If the $L_{D-Path}$ maintains the value of the shortest path length for the whole evaluation period, one could deduce that the T-Point has already occurred. Thus, the recorded running time in Step ii is the running time used for comparison; otherwise, return to Step ii.	
\end{enumerate}

\subsubsection{Comparing different accelerated PPAs using the proposed method}
\label{sec.4.3.2}
\quad \\
\noindent\textbf{Data Set 3:} The instances `Da-Com-1' to `Da-Com-8' mentioned in Data Set 1 are adopted.

\noindent\textbf{Tested algorithms:} The OPPA, the OPPA-D, the EHPA and the APS are evaluated. When tested, each algorithm would be run 30 times in each graph. 

\begin{table}
	\centering
	\caption{Average running time of tested algorithms in Data Set 3}
	\resizebox{0.5\textwidth}{16mm}{
		\begin{tabular}{ccccc}
			\toprule[0.75pt]
			\multirow{2}{*}{Instacnes} 	& \multicolumn{4}{c}{Running time (seconds)}\\					
			&	OPPA&	EHPA&	APS&	OPPA-D (Ours)	\\\hline
			Da-Com-1	&	0.0006&	0.0025&	0.0029&	0.0006	\\
			Da-Com-2	&	0.0068&	0.0038&	0.0065&	0.0046	\\
			Da-Com-3	&	0.0017&	0.0040&	0.0346&	0.0020	\\
			Da-Com-4	&	0.1194&	0.2675&	0.2061&	0.1213	\\
			Da-Com-5	&	0.2778&	0.6585&	0.5086&	0.2833	\\
			Da-Com-6	&	0.4198&	1.1887&	1.0730&	0.4267	\\
			Da-Com-7	&	1.5913&	11.5127&	15.3770&	1.5929	\\
			Da-Com-8	&	15.2210&	216.5146&	262.6555&	15.3558	\\
			\bottomrule[0.75pt] 
		\end{tabular}
	}
	\\\footnotesize{$^a$Values presented in the table are the average value of 30 trials.}
	\label{table.exp3_time}
\end{table}

\begin{table}
	\centering
	\caption{Average number of iterations of tested algorithms in Data Set 3}
	\resizebox{0.5\textwidth}{16mm}{
		\begin{tabular}{ccccc}
			\toprule[0.75pt]
			\multirow{2}{*}{Instacnes} 	& \multicolumn{4}{c}{Number of iterations}\\					
			&	OPPA&	EHPA&	APS&	OPPA-D (Ours)	\\\hline
			Da-Com-1	&	2&	2&	16&	2	\\
			Da-Com-2	&	9&	9&	16&	9	\\
			Da-Com-3	&	1&	1&	16&	1	\\
			Da-Com-4	&	15&	15&	16&	15	\\
			Da-Com-5	&	15&	15&	16&	15	\\
			Da-Com-6	&	12&	12&	16&	12	\\
			Da-Com-7	&	9&	9&	16&	9	\\
			Da-Com-8	&	13&	13&	16&	13	\\	
			\bottomrule[0.75pt] 
		\end{tabular}
	}
	\\\footnotesize{$^a$Values presented in the table are the average value of 30 trials.}
	\label{table.exp3_ite}
\end{table}

Tables \ref{table.exp3_time} and \ref{table.exp3_ite} presents the experimental results. It is not a surprise that the running time and the number of iterations of the OPPA-D and the OPPA are approximately the same, as the difference between the OPPA and the OPPA-D lies only in the termination criterion. The similar performance of the OPPA and the OPPA-D under the proposed evaluation method proves that the method developed by us could exclude the effects brought by different termination criteria. 

Though the EHPA share the same number of iterations with the OPPA and the OPPA-D, the running time of EHPA is approximately 14 times longer than that of the OPPA and the OPPA-D in instance `Da-Com-8'. The above findings indicate that when the effects of the termination criterion is excluded, the EHPA fails to accelerate the PPA as expected; instead, the running time per iteration is significantly longer.

According to Table \ref{table.exp3_time}, on average, the running time of the APS is even longer than the EHPA. Furthermore, it is surprising that, in all instances, the number of the iterations of the APS is the same, which is 16. We carefully examined the codes provided by the authors \cite{GG_APS} and found out that at the beginning of the iterative procedure of the APS, the algorithm has a 15-iteration initialization run. One could easily deduce that the APS only uses one iteration to reach the T-Point after the initialization. Hence, it suggests that the initialization of the APS costs the majority of the running time of the algorithm, and further leads to the inefficiency of the APS.

In general, the proposed methods do have the ability to avoid the effects brought by different stopping criteria, as the performance of the OPPA and the OPPA-D is approximately the same in the experiment. The most important finding in this experiment is that the accelerated OPPAs, i.e., the EHPA and the APS fail to accelerate the PPA as expected. A potential reason might be the failure of the authors to not consider the possible effects brought about by the termination criteria.

\section{Discussions \& Conclusions}
\label{sec.5}
Though many studies have proposed various solutions to accelerate the PPA, few have considered the core question: when does the physarum solver distinguish the shortest path from other paths? By defining the concepts of the T-Point and the D-Path in Section \ref{sec.3}, we showed that the physarum solver distinguishes the shortest path at the T-Point. We also propose the use of the T-Point as the inherent limit, beyond which the algorithm can terminate, hence improving efficiency while maintaining accuracy. The mathematical proof remains in the pipelines for further work. We hope that in the future, the phenomenon of T-Point could also be experimentally confirmed in biological experiments.

Based on the T-Point and the D-Path, a new termination criterion for the OPPA and a corresponding algorithm, OPPA-D, are developed. The OPPA-D is actually the OPPA with the newly proposed termination criterion. From our experiments in Section \ref{sec.4.1}, the OPPA-D outperforms two state-of-the-art and widely used accelerated OPPAs. In order words, the OPPA significantly outperforms the accelerated OPPAs despite a slight modification to the stopping criterion. This begs another question: does the accelerated OPPAs beat the OPPA as claimed?

In section \ref{sec.4.2}, we found that although the proposed termination criterion is superior to the traditional one, both of them are sensitive to the predefined parameters, which would further affect the validity of comparing different accelerated OPPAs. Hence, in order to evaluate the efficiency of accelerated OPPAs, the effects of the termination criteria need to be excluded. Thus, we proposed a new evaluation method to compare the accelerated OPPAs more objectively in Section \ref{sec.4.3.1}. From our results in section \ref{sec.4.3.2} one finds that the accelerated OPPAs is not as efficient as the OPPA, as previously reported. Previous research did not evaluate the real performance of the accelerated OPPA by neglecting the possible effects brought by the termination criteria. However, with the proposed T-Point and the D-Path, and the developed evaluation method, experiments in our studies demonstrate that the OPPA is the fastest algorithm among the tested algorithms. We hope the findings in this paper would change the standing critique that OPPAs are not efficient.

\section*{Acknowledgment}
The work is partially supported by National Natural Science Foundation of China (Grant No. 61973332).

\ifCLASSOPTIONcaptionsoff
  \newpage
\fi



\bibliographystyle{IEEEtran}
\bibliography{DPPA}




\end{document}